\documentclass{article}
\usepackage [utf8]{inputenc}

\usepackage{microtype}
\usepackage{graphicx}
\usepackage{subfigure}
\usepackage{booktabs} 
\usepackage{enumitem}       
\usepackage{multirow}
\usepackage[toc,page,header]{appendix}
\usepackage[nottoc]{tocbibind}
\usepackage[symbol]{footmisc}

\usepackage{mathtools}
\DeclarePairedDelimiter\ceil{\lceil}{\rceil}
\DeclarePairedDelimiter\floor{\lfloor}{\rfloor}

\usepackage{hyperref}
\usepackage{url}

\usepackage{listings}
\usepackage{authblk}

\usepackage[accepted]{icml2024}

\usepackage{amsmath}
\usepackage{amssymb}
\usepackage{mathtools}
\usepackage{amsthm}
\usepackage{adjustbox}

\usepackage[capitalize,noabbrev]{cleveref}
\Crefname{section}{\mbox{\S\hspace*{-0.25ex}}}{\mbox{\S\hspace*{-0.25ex}}}
\Crefname{equation}{Eq.}{Eqs.}
\Crefname{figure}{Fig.}{Figs.}
\Crefname{table}{Tab.}{Tabs.}
\Crefname{appendix}{\S$\!$}{\S$\!$}

\usepackage{caption}
\usepackage{enumitem} 
\setlist[itemize]{noitemsep,topsep=0ex}    
\setlist{leftmargin=3mm}

\usepackage[light,scaled=0.85]{roboto-mono}

\theoremstyle{plain}

\theoremstyle{definition}

\theoremstyle{remark}

\usepackage[textsize=tiny,colorinlistoftodos]{todonotes}

\usepackage{xspace}

\newcommand{\model}{\texttt{Lag-Llama}\xspace}

\definecolor{darkbrown}{rgb}{0.43, 0.21, 0.1}
\usepackage{color, colortbl}
\icmltitlerunning{\model}

\newif\ifcomments

\commentstrue

\ifcomments
\newcommand{\comments}[1]{#1}
\else
\newcommand{\comments}[1]{}
\fi

\definecolor{ao}{rgb}{0.0, 0.5, 0.0}
\definecolor{darkred}{rgb}{0.55, 0.0, 0.0}

\begin{document}

\twocolumn[
\icmltitle{\model: Towards Foundation Models for Probabilistic~Time~Series~Forecasting}



\icmlsetsymbol{equal}{*}

\begin{icmlauthorlist}

\icmlauthor{\textsuperscript{\normalsize \textbf{*}}Kashif Rasul}{1}
\icmlauthor{\textsuperscript{\normalsize \textbf{*}}Arjun Ashok}{2,3,4}\\ \vspace{0.2em}
\icmlauthor{$^\diamondsuit$Andrew Robert Williams}{3,4}
\icmlauthor{$^\diamondsuit$Hena Ghonia}{3,4}
\icmlauthor{Rishika Bhagwatkar}{3,4}\\ \vspace{0.2em}
\icmlauthor{$^\spadesuit$Arian Khorasani}{3,4}
\icmlauthor{$^\spadesuit$Mohammad Javad Darvishi Bayazi}{3}
\icmlauthor{$^\spadesuit$George Adamopoulos}{5}
\icmlauthor{$^\spadesuit$Roland Riachi}{3,4}\\ \vspace{0.2em}
\icmlauthor{Nadhir Hassen}{3,4}
\icmlauthor{Marin Biloš}{1}
\icmlauthor{$^\triangle$Sahil Garg}{1}
\icmlauthor{$^\triangle$Anderson Schneider}{1}
\icmlauthor{$^\triangle$Nicolas Chapados}{2,4}\\ \vspace{0.2em}
\icmlauthor{$^\triangle$Alexandre Drouin}{2,4}
\icmlauthor{$^\triangle$Valentina Zantedeschi}{2}
\icmlauthor{$^\clubsuit$Yuriy Nevmyvaka}{1}
\icmlauthor{$^\clubsuit$Irina Rish }{3,4}

\end{icmlauthorlist}

\icmlaffiliation{1}{Morgan Stanley, New York, USA}
\icmlaffiliation{2}{ServiceNow Research, Montr\'eal, Canada }
\icmlaffiliation{3}{Universit\'e de Montr\'eal, Montr\'eal, Canada }
\icmlaffiliation{4}{Mila-Quebec AI Institute, Montr\'eal, Canada}
\icmlaffiliation{5}{McGill University, Montr\'eal, Canada}

\icmlcorrespondingauthor{Arjun Ashok}{arjun.ashok@servicenow.com}
\icmlcorrespondingauthor{Kashif Rasul}{kashif.rasul@gmail.com}

\vskip 0.3in
]



\printAffiliationsAndNotice{\textsuperscript{\normalsize \textbf{*}}Co-first authorship, authors contributed equally, order arbitrary. $^\diamondsuit$$^\spadesuit$$^\triangle$$^\clubsuit$ Authors in each group contributed equally, order arbitrary.} 

\begin{abstract}
Over the past years, foundation models have caused a paradigm shift in machine learning due to their unprecedented capabilities for zero-shot and few-shot generalization. However, despite the success of foundation models in modalities such as natural language processing and computer vision, the development of foundation models for time series forecasting has lagged behind. We present Lag-Llama, a general-purpose foundation model for univariate probabilistic time series forecasting based on a decoder-only transformer architecture that uses lags as covariates. Lag-Llama is pretrained on a large corpus of diverse time series data from several domains, and demonstrates strong zero-shot generalization capabilities compared to a wide range of forecasting models on downstream datasets across domains. Moreover, when fine-tuned on relatively small fractions of such previously unseen datasets, Lag-Llama achieves state-of-the-art performance, outperforming prior deep learning approaches, emerging as the best general-purpose model on average. Lag-Llama serves as a strong contender to the current state-of-art in time series forecasting and paves the way for future advancements in foundation models tailored to time series data.
\end{abstract}

\section{Introduction}

Probabilistic time series forecasting is an important practical problem arising in a wide range of applications, from finance and weather forecasting to brain imaging and computer systems performance management  \cite{peterson2017decision}.  Accurate probabilistic forecasting is usually an essential step towards the subsequent decision-making in such practical domains. The probabilistic nature of such forecasting endows decision-makers with a notion of uncertainty, allowing them to consider a variety of future scenarios, along with their respective likelihoods.
Various methods have been proposed for this task, ranging from classical autoregressive models \cite{hyndman2021forecasting} to the more recent neural forecasting methods based on deep learning architectures  \cite{torres2021deep}. Note that the overwhelming majority of these previous approaches are focused on building dataset-specific models, i.e. models tested on the same dataset in which training is performed.

Recently, however, machine learning is witnessing a paradigm shift due to the rise of {\em foundation models} \cite{bommasani2022opportunities} --- large-scale, general-purpose neural networks pretrained in an unsupervised manner on large amounts of diverse data across various data distributions. Such models demonstrate remarkable few-shot generalization capabilities on a wide range of downstream datasets \cite{NEURIPS2020_1457c0d6}, often outperforming dataset-specific models. Following the successes of foundation models in language and image processing domains\cite{openai2023gpt4,radford2021learningCLIP}, we aim to develop foundation models for time series, investigate their behaviour at scale, and push the limits of transfer achievable across diverse time series domains.

In this paper, we present \model --- a foundation model for probabilistic time series forecasting trained on a large collection of open time series data, and evaluated on unseen time series datasets.
%
We investigate the performance of \model across several settings where unseen time series datasets are encountered downstream with different levels of data history being available, and show that \model performs comparably or better against state-of-the-art dataset-specific models.

\textbf{Our contributions:}
\begin{itemize}[leftmargin=2ex]
    \item We present \model, a foundation model for univariate probabilistic time series forecasting based on a simple decoder-only transformer architecture that uses lags as covariates.
    \item We show that \model, when pretrained from scratch on a broad, diverse corpus of datasets, has strong \textit{zero-shot} performance on unseen datasets, and performs comparably to models trained on the specific datasets.
    \item \model also demonstrates \textit{state-of-the-art} performance across diverse datasets from different domains after \textit{finetuning}, and emerges as the best general-purpose model without any knowledge of downstream datasets.
    \item We demonstrate the strong \textit{few-shot adaptation} performance of \model on previously unseen datasets, across varying fractions of data history being available.
    \item We investigate the diversity of the pretraining corpus used to train \model, and present the scaling laws of \model with respect to the pretraining data. 
\end{itemize}
















\section{Related Work}

\textbf{\emph{Statistical models}} have been the cornerstone of time series forecasting for decades, evolving continuously to address complex forecasting challenges. Traditional models such as ARIMA (Autoregressive Integrated Moving Average) set the foundation by using autocorrelation to forecast future values. 
ETS (Error, Trend, Seasonality) models advanced this by decomposing a time series into its fundamental components, allowing for more nuanced forecasting that captures trends and seasonal patterns. Theta models, introduced by \citet{ASSIMAKOPOULOS2000521}, represented another significant advancement in time series forecasting. By applying a decomposition technique combining both long-term trend and seasonality, these models offer a simple yet effective method for forecasting
Despite the success of the considerable successes of these statistical models and more advanced ones \citep{6979ac51-118f-3133-9fc7-62cf907a4de3, SYNTETOS2005303, 3b1355aedd1041f1853e609a410576f3}, these models share common limitations. Their primary shortfall lies in their inherent assumption of linear relationships and stationarity in time series data, which is often not the case in real-world scenarios marked by abrupt changes and non-linear dynamics. Furthermore, they may require extensive manual tuning and domain knowledge to select appropriate models and parameters for specific forecasting tasks.

 

\textbf{\emph{Neural forecasting} }is a rapidly developing research area following the explosion of machine learning \citep{1Benidis_2022}.
Various architectures have been developed for this setting, starting with RNN-based and LSTM-based models \citep{salinas2020deepar, wen2018multihorizon}.
%
%
More recently in light of the recent success of transformers \citep{NIPS2017_3f5ee243} for sequence-to-sequence modelling for natural language processing, many variations of transformers have been proposed for time series forecasting.
Different models \citep{Yuqietal-2023-PatchTST, wu2020deep, NEURIPS2020_c6b8c8d7}  process the input time series in different ways to be digestible by a vanilla transformer, then re-process the output of a transformer for a point forecast or a probabilistic forecast.
On the other hand, various other works propose alternative strategies to vanilla attention and build off the transformer architecture, for better models tailored for time series \citep{LIM20211748, li2023timae, ashok2023tactis2, Oreshkin2020N-BEATS:, Zhou_Zhang_Peng_Zhang_Li_Xiong_Zhang_2021, wu2021autoformer, woo2023etsformer, liu2022nonstationary, pmlr-v162-zhou22g, liu2022pyraformer, ni2023basisformer, NEURIPS2019_6775a063, gulati2020conformer}.

\textbf{\emph{Foundation models}} are an emerging paradigm of self-supervised (or) unsupervised learning on large datasets \citep{bommasani2022opportunities}.
Many such models \citep{devlin-etal-2019-bert, openai2023gpt4, chowdhery2022palm, radford2021learningCLIP, wang2022imageBEIT} have demonstrated adaptability across modalities, extending beyond web data to scientific domains such as protein design \citep{protein22}.
Scaling the model, dataset size and data diversity have also been shown to result in remarkable transfer capabilities and excellent few-shot learning on novel datasets and tasks \citep{LEARN10.5555/296635.296639, brown2020language}.
Self-supervised learning techniques have also been proposed for time series \citep{li2023timae, woo2022cost, yeh2023foundation}.
Most related to our work is \citet{yeh2023foundation} who train on a corpus of time series datasets. The key difference is that they validate their model only on the downstream classification tasks, and do not validate on forecasting tasks.
Works such as \texttt{Time-LLM} \citep{jin2023timellm}, \texttt{LLM4TS} \citep{chang2023llm4ts}, \texttt{GPT2(6)} \citep{zhou2023fits}, \texttt{UniTime} \citep{liu2023unitime}, and \texttt{TEMPO} \citep{anonymous2024tempo} freeze LLM encoder backbones while simultaneously fine-tuning/adapting the input and distribution heads for forecasting.
The main goal of our work is to apply the foundation model approach to time series data and to investigate the extent of the transfer achievable across a wide range of time series domains.

\section{Probabilistic Time Series Forecasting}
\label{problem}
We consider a dataset of \smash{$D \geq 1$} univariate  time series,  $\mathcal{D}_{\mathrm{train}} = \{x_{1:T^i}^i\}_{i=1}^D$ sampled at a specific  discrete set of  time  points $t \in \{1, \ldots, T^i\}$ where $T^i$ represents the length of the time series $i$.  
Given this  dataset, we aim to train a predictive model that can accurately predict the  values at the future $P \geq 1$   time points; we refer to these timesteps of our  $D$ time series as to the {\em test dataset}, denoted  \smash{$\mathcal{D}_{\mathrm{test}} = \{x_{T^{i}+1:T^i+P}^i\}_{i=1}^D$}. 
%

The  \emph{univariate} probabilistic time series forecasting problem  involves modelling an  unknown joint distribution of the $P$ future values of a one-dimensional sequence given its observed past until timestep $t$ from which prediction should be performed, and covariates:
\begin{equation}
    p_{\mathcal{\phi}}(x_{t+1:t+P}^i \mid x_{1:t}^i, \mathbf{c}^i_{1:t+P}).
\end{equation}
%
where $\mathcal{\phi}$ represents the parameters of a parametric distribution.
In practice, rather than considering the whole history of each time series $i$, which can vary considerably, we can instead sub-sample fixed context windows of size $C \geq 1$ of our choosing from the complete time series and learn an approximation of the unknown distribution of the next $P$ future values given  the covariates: 
\begin{equation}\label{eqn:dist}
    p_{\mathcal{\phi}}(x_{C+1:C+P}^i \mid x_{1:C}^i, \mathbf{c}^i_{1:C+P}).
\end{equation}
When the distribution is modeled by a neural network with parameters $\theta$, predictions are then conditioned on these (learned) parameters $\theta$.
We  will approximate the distribution in \cref{eqn:dist} by  an autoregressive model, using the chain rule of probability as follows:
\begin{multline*}
    p_{\mathcal{\phi}}(x_{C+1:C+P}^i \mid x_{1:C}^i, \mathbf{c}^i_{1:C+P}; \theta)  =  
    \\ \prod_{t=C+1}^{C +P} p_{\mathcal{\phi}}(x_{t}^i \mid x_{1:t-1}^i, \mathbf{c}^i_{1:t-1}; \theta).
\end{multline*}


\section{\model}
We present \model, a {\em foundation model} for univariate probabilistic forecasting. The first step in building such a foundation model for time series is training on a large corpus of diverse time series. When training on heterogenous univariate time series corpora, the frequency of the time series in our corpus varies. Further, when adapting our foundation model to downstream datasets, we may encounter new frequencies and combinations of seen frequencies, which our model should be capable of handling. We now present a general method for tokenizing series from such a dataset, without directly relying on the frequency of any specific dataset, and thus potentially allowing unseen frequencies and combinations of seen frequencies to be used at test time.

\subsection{Tokenization: Lag Features}\label{sec:lagFeatures}

\begin{figure}[ht]
\begin{center}
\centerline{\includegraphics[width=\columnwidth]{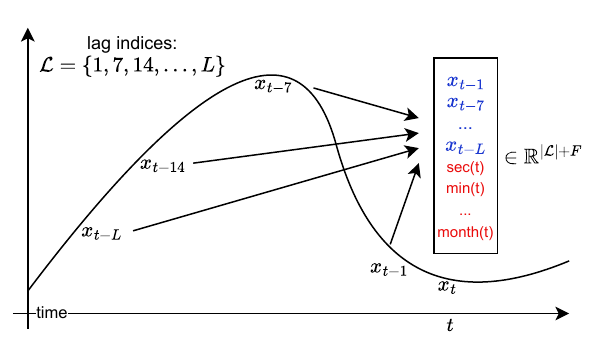}}
\caption{For a time series, we depict the tokenization at the timestep $t$ of the value $x_t$ which contains lag features constructed using an example set of lag indices $\mathcal{L}$, where each value in the vector is from the \emph{past} of $x_t$ (in blue), and $F$ possible temporal covariates (date-time features) constructed from timestamp $t$ (red).
%
%
}
\label{fig:lag-schematics}
\end{center}
\vskip -0.2in
\end{figure}

The tokenization scheme of \model involves constructing \textit{lagged features} from the prior values of the time series, constructed according to a specified set of appropriate lag indices that include quarterly, monthly, weekly, daily, hourly, and second-level frequencies. Given a sorted set of positive lag indices $ \mathcal{L} = \{1, \ldots, L \}$\footnote{Note that $\mathcal{L}$ refers to the  list of lag indices, while $L$ is the last lag index in the sorted list $\mathcal{L}$},  we define the lag operation on a particular time value as $x_t  \mapsto  \mathbf{k}_{t} \in \mathbb{R}^{|\mathcal{L}|}$  where each entry $j$ of  $\mathbf{k}_{t}$ is given by $\mathbf{k}_t[j] = x_{t - \mathcal{L}[j]}$. Thus to create lag features for some context-length window $x_{1:C}$ we need to sample a larger window with $L$ more historical points denoted by  $x_{-L:C}$\footnote[5]{This is since a history of $L$ points in time is needed for all points in the context, starting from the first point in the context}. In addition to these lagged features, we add date-time features of \emph{all} the frequencies in our corpus, namely second-of-minute, hour-of-day, etc. up till the quarter-of-year from the time index $t$. Note that while the primary goal of these date-time features is to provide additional information, for any time series, all except one date-time feature will remain constant from one time-step to the next, and from the model can implicitly make sense of the frequency of the time series as well. Assuming we employ a total of $F$ date-time features, each of our tokens is of size $|\mathcal{L}|+ F$. \cref{fig:lag-schematics} shows an example tokenization. We note that a downside to using lagged features in tokenization is that it requires an $L$-sized or larger context window. 

\subsection{\model Architecture} \label{sec:lagllama}

\model's architecture is based on the decoder-only transformer-based architecture LLaMA \citep{touvron2023llama}.  

\begin{figure}[ht]
\vskip 0.2in
\begin{center}
\centerline{\includegraphics[width=\columnwidth]{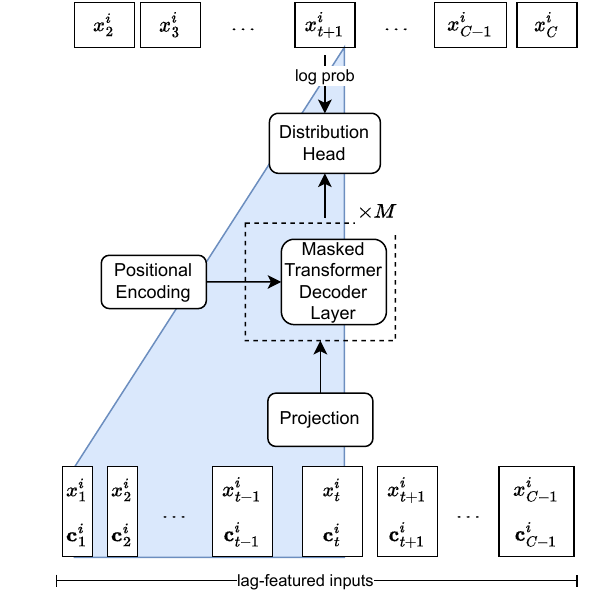}}
\caption{The \model architecture. \model learns to output a distribution over the values of the next time step based on lagged input features.
    The input to the model is the token of a univariate time series $i$ at a given timestep, $\mathbf{x}^i_t$, constructed as described in Sec.\ref{sec:lagFeatures}. Here, we use $\mathbf{c}_t^i$ to refer to all additional covariates used along with the value at a timestep $t$, which include the $|\mathcal{L}|$ lags, $F$ date-time features, and summary statistics.
    The inputs are projected through $M$ masked decoder layers.
    The features are then passed through the distribution head and trained to predict the parameters of the forecast distribution of the next timestep.
    }
\label{fig:lag-llama}
\end{center}
\vskip -0.2in
\end{figure}

Fig. \ref{fig:lag-llama} shows a general schematic of this model with $M$ decoder layers. A univariate sequence of length \smash{$x^i_{-L:C}$} 
along with its covariates is tokenized by concatenating the covariate vectors to a sequence of $C$ tokens $\mathbf{x}^i_{1:C}$.
These tokens are passed through a shared linear projection layer that maps the features to the hidden dimension of the attention module. Similar to in \citet{touvron2023llama}, \model incorporates pre-normalization via the RMSNorm \citep{NEURIPS2019_1e8a1942} and Rotary Positional Encoding  (RoPE) \citep{su2021roformer} at 
each attention layer's query and key representations as in LLaMA \citep{touvron2023llama}.

After passing through the causally masked transformer layers, the model predicts the parameters $\phi$ of the forecast distribution of the \textit{next timestep}, where the parameters are output by a \textit{parametric distribution head}, as described in Sec. \ref{sec:distributionHead}. The negative log-likelihood of the predicted distribution of all predicted timesteps is minimized.

At inference time, given a time series of size at least $L$, 
 we can construct a feature vector 
that is passed to the model to obtain the distribution of the next time point. 
In this fashion, via \emph{greedy} autoregressive decoding, we can obtain many simulated trajectories of the future up to our chosen prediction horizon $P \geq 1$.
From these empirical samples, we can calculate the uncertainty intervals for downstream decision-making tasks and metrics with respect to held-out data.

\subsection{Choice of Distribution Head}\label{sec:distributionHead}

The last layer of \model is a distinct layer known as the \emph{distribution head}, which projects the model's features to the parameters of a probability distribution.
We can combine different distribution heads with the representational capacity of the model to output the parameters $\phi$ of any parametric probability distribution. 
For our experiments, we adopt a Student's $t$-distribution \citep{student1908probable} and output the three parameters corresponding to this distribution, namely its degrees of freedom, mean, and scale, with appropriate non-linearities to ensure the appropriate parameters stay positive.
More expressive choices of distributions, such as normalizing flows \citep{rasul2021multivariate} and copulas \citep{SalinasGPVar, Drouin2022TACTiSTC, ashok2023tactis2} are potential choices of distribution heads, however with the potential overhead of difficulties in model training and optimization. The goal of our work was to keep the model as simple as possible, which led us to adopt a simple parametric distributional head. We leave the exploration of such distribution heads for future work.

\subsection{Value Scaling}\label{sec:valueScaling}

When training on a large corpus of time series data from different datasets and domains, each time series can be of different numerical magnitude. Since we pretrain a foundation model over such data, we utilize the scaling heuristic \citep{DBLP:journals/corr/FlunkertSG17} where for each univariate window, we calculate its mean value \smash{$\mu^i = \sum_{t=1}^{C} x^i_t/C$}  and variance \smash{$\sigma^i$}. 
We can then replace the time series \smash{$x^i_{1:C}$} in the window by \smash{$\{(x^i_t - \mu^i) / \sigma^i \}_{t=1}^{C}$}. 
We also incorporate $\mu^i$ and $\sigma^i$ as time independent real-valued covariates for each token, to give the model information of the statistics of the inputs, which we call \textit{summary statistics}.

During training and obtaining likelihood, the values are transformed using the mean and variance, while sampling, every timestep of data that is sampled is de-standardized using the same mean and variance. In practice, instead of the standard scaler, we find the following standardization strategy works well when pretraining our model.
%

\textbf{Robust Standardization} ensures that our time series processing is robust to outliers. This procedures normalizes the series by removing the median and scaling according to the Interquartile Range (IQR) \cite{dekking2005modern}.
For a context-window sized series $x_{1:C} = \{x_1, x_2, ..., x_C \}$ we standardize each time point as:
\begin{align}
    x^{\prime}_t &=\frac{x_t -\operatorname{Med}(x_{1:C})}{\operatorname{IQR}(x_{1:C})}, \quad \textrm{where} \\
    \operatorname{IQR}(x_{1:C}) &= \operatorname{Med}(\{x_{\ceil{C/2}:C}\}) - \operatorname{Med}(\{x_{1:\floor{C/2}}\}).
\end{align}

\subsection{Training Strategies} \label{sec:aug}

We employ a series of training strategies to effectively pretrain \model on the corpus of datasets. Firstly, we find that employing a stratified sampling approach where the datasets in the corpus are weighed by the amount of \emph{total} number of series is useful when sampling random windows from the pretraining corpus. Further, we find that employing time series augmentation techniques of \texttt{Freq-Mix} and \texttt{Freq-Mask} \citep{chen2023fraug} serve useful to reduce overfitting. We search the hyperparameters of these augmentation strategies as part of our hyperparameter search.

\section{Experimental Setup}



\subsection{Datasets}
We collate a diverse corpus of $27$ time series datasets from several sources across six different semantically grouped domains such as \textbf{energy}, \textbf{transportation}, \textbf{economics}, \textbf{nature}, \textbf{air quality} and \textbf{cloud operations}; each dataset has a different set of  characteristics, such as prediction lengths, number of series, lengths of each series, and frequencies. 

We leave out a few datasets from each domain for testing the few-shot generalization abilities of the pretrained model,   whle using the remaining datasets for pretraining the foundation model. Furthermore, we set aside datasets from entirely different domains to assess our model's performance on data that may lack any potential similarity to the datasets in pretraining. Such a setup mimics the real-world use of our model, where one may adapt it for datasets that fall closely within the distribution of domains that the model has been trained on, as well as datasets in completely different domains. Our pretraining corpus comprises a total of $7,965$ different univariate time series, each of different lengths, when put together, comprising a total of around $352$ million data windows (tokens) for our model to train on. App. \cref{app:dataset-details} lists the datasets we use, along with their sources and properties, their respective domains, and the dataset split used in our experiments.

Note that the term ``domain'' used here is just a \textit{label} used to  group several datasets, which does not represent a common source or data distribution;  each of the pretraining and test datasets  possesses very different general characteristics (patterns, seasonalities), apart from having other distinct properties. We use the default prediction length of each dataset for evaluation and ensure that there is a wide variety of prediction horizons in our unseen corpus of datasets, to evaluate models on short-term, medium-term, and long-term forecasting setups. App. \cref{app:dataset-details} lists the different datasets used in this work, along with the sources and properties of each dataset. Sec. \cref{sec:dataset-analysis} analyses the diversity of our corpus of datasets.





\begin{table*}[t]

\caption{\textit{CRPS of \model zero-shot and on finetuning on the {unseen datasets}, compared to supervised baselines trained solely on the respective datasets. Lower is better. A mean or standard deviation of 0.0000 signifies that the first non-zero digit is beyond $3$ decimal places. The best results are in \textbf{bold}, and the second best results are in \textcolor{darkbrown}{brown}.}} 
\label{table:performance}
\vskip 0.15in
\begin{center}
\begin{scriptsize}
\begin{sc}
\addtolength{\tabcolsep}{-0.6em}
\begin{tabular}{lcccccccc}
\toprule
\multirow{2}{*}{\textbf{Model}} & \multicolumn{7}{c}{\textbf{Dataset}} & \multirow{2}{*}{\textbf{Average Rank}} \\
\cmidrule(lr){2-8}
& \texttt{weather} & \texttt{ped-counts} & \texttt{ett-m2} & \texttt{platform-delay} & \texttt{requests} & \texttt{beijing-pm2.5} & \texttt{exchange} & \\

\midrule
\textbf{Supervised} & & & & & & & \\
\midrule

ETSFormer & 0.528$\pm$0.175 & 0.275$\pm$0.024 & 0.140$\pm$0.002 & 0.171$\pm$0.025 & 0.218$\pm$0.070 & 0.266$\pm$0.099 & 0.029$\pm$0.014 & 13.000 \\
NPTS & 0.276$\pm$0.000 & 0.684$\pm$0.006 & 0.139$\pm$0.000 & 0.132$\pm$0.001 & 0.085$\pm$0.001 & 0.170$\pm$0.003 & 0.059$\pm$0.001 & 12.714 \\
OFA & 0.265$\pm$0.006 & 0.605$\pm$0.023 & 0.130$\pm$0.006 & 0.213$\pm$0.011 & 0.121$\pm$0.011 & 0.130$\pm$0.009 & 0.015$\pm$0.001 & 11.357 \\
AutoFormer & 0.240$\pm$0.021 & 0.247$\pm$0.011 & 0.088$\pm$0.014 & 0.152$\pm$0.030 & 0.301$\pm$0.178 & 0.151$\pm$0.002 & 0.037$\pm$0.025 & 11.000 \\
CrostonSBA & 0.177$\pm$0.000 & 0.594$\pm$0.000 & 0.102$\pm$0.000 & 0.097$\pm$0.000 & 0.042$\pm$0.000 & 0.198$\pm$0.000 & 0.031$\pm$0.000 & 9.429 \\
AutoARIMA & 0.213$\pm$0.000 & 0.755$\pm$0.000 & nan$\pm$nan & 0.112$\pm$0.000 & 0.076$\pm$0.000 & 0.110$\pm$0.000 & \textcolor{darkbrown}{0.009$\pm$0.000} & 8.333 \\
AutoETS & 0.215$\pm$0.000 & 0.625$\pm$0.000 & 0.081$\pm$0.000 & 0.297$\pm$0.000 & 0.041$\pm$0.000 & 0.090$\pm$0.000 & \textbf{0.008$\pm$0.000} & 8.000 \\
DynOptTheta & 0.217$\pm$0.000 & 1.817$\pm$0.000 & 0.049$\pm$0.000 & 0.118$\pm$0.000 & 0.055$\pm$0.000 & 0.108$\pm$0.000 & \textbf{0.008$\pm$0.000} & 7.857 \\
Informer & 0.172$\pm$0.011 &\textbf{ 0.223$\pm$0.005} & 0.070$\pm$0.003 & 0.106$\pm$0.009 & 0.104$\pm$0.012 & \textbf{0.057$\pm$0.003} & 0.017$\pm$0.004 & 6.429 \\
DeepAR & 0.148$\pm$0.004 & 0.239$\pm$0.002 & 0.068$\pm$0.003 & \textbf{0.068$\pm$0.003} & 0.045$\pm$0.009 & 0.154$\pm$0.000 & 0.012$\pm$0.000 & 5.714 \\
PatchTST & 0.178$\pm$0.013 & 0.254$\pm$0.001 & 0.035$\pm$0.000 & 0.094$\pm$0.001 & 0.024$\pm$0.003 & 0.145$\pm$0.001 & 0.011$\pm$0.000 & 5.643 \\
N-BEATS & \textcolor{darkbrown}{0.134$\pm$0.003} & 0.267$\pm$0.018 & 0.031$\pm$0.005 & 0.112$\pm$0.007 & 0.021$\pm$0.005 & \textcolor{darkbrown}{0.081$\pm$0.004} & 0.024$\pm$0.004 & 5.071 \\
TFT & 0.151$\pm$0.016 & 0.268$\pm$0.009 & \textcolor{darkbrown}{0.030$\pm$0.000} & 0.099$\pm$0.001 & \textcolor{darkbrown}{0.015$\pm$0.003} & 0.156$\pm$0.000 & \textbf{0.008$\pm$0.000 }& \textcolor{darkbrown}{5.000} \\

\midrule
\textbf{Zero-shot} & & & & & & & \\
\midrule
\model & 0.164$\pm$0.001 & 0.285$\pm$0.033& 0.063$\pm$0.002 & \textcolor{darkbrown}{0.091$\pm$0.002} & 0.090$\pm$0.015 & 0.130$\pm$0.009 & 0.011$\pm$0.001 & 6.714\\
\midrule
\textbf{Finetuned} & & & & & & & \\
\midrule
\model & \textbf{0.132$\pm$0.001} & \textcolor{darkbrown}{0.227$\pm$0.010}  & \textbf{0.017$\pm$0.001} & 0.096$\pm$0.002 & \textbf{0.012$\pm$0.002} & 0.125$\pm$0.021 & \textcolor{darkbrown}{0.009$\pm$0.000} & \textbf{2.786}\\
\bottomrule
\end{tabular}
\end{sc}
\end{scriptsize}
\end{center}
\vskip -0.1in
\end{table*}

\subsection{Baselines}


We compare the performance of \model to that of a large set of baselines, including both standard statistical models, as well as deep neural networks. 
    
Through AutoGluon \citep{shchur2023autogluontimeseries}, an AutoML framework for probabilistic time series forecasting, we benchmark against $5$ well-known statistical time series forecasting models: AutoARIMA \citep{JSSv027i03} and AutoETS \citep{JSSv027i03} which are established statistical models that tune model parameters locally for each time series \citep{JSSv027i03}, CrostonSBA (Syntetos and Boylan Approximate) \citep{6979ac51-118f-3133-9fc7-62cf907a4de3, SYNTETOS2005303} an intermittent demand forecasting model using Croston’s model with the Syntetos-Boylan bias correction approach, DynOptTheta (The Dynamically Optimized Theta model) \citep{box1976time} a statistical forecasting method that is based on the decomposition of the time series into trend, seasonality and noise, and NPTS (Non-Parametric Time Series Forecaster) \citep{shchur2023autogluontimeseries}, a local forecasting method that assumes a non-parametric sampling distribution. We further compare with $3$ strong deep-learning methods through the same AutoGluon framework: DeepAR \citep{salinas2020deepar}, an autoregressive RNN-based method that has been shown to be a strong contender for probabilistic forecasting \citep{gluonts_jmlr}, PatchTST \citep{nie2023a} a univariate transformer-based method that uses patching to tokenize time series, TFT (Temporal Fusion Transformer) \citep{LIM20211748}, an attention-based architecture with recurrent and feature-selection layers. 
    
We benchmark against $4$ more deep learning models: N-BEATS \citep{oreshkin2020nbeats}, a neural network architecture that uses a recursive decomposition based on projecting residual signals on learned basis functions, Informer \citep{zhou2021informer}, an efficient autoregressive transformer-based method that uses a ProbSparse self-attention mechanism to handle extremely long sequences, AutoFormer \citep{wu2022autoformer}, a transformer-based architecture with an Auto-Correlation mechanism based on the series periodicity, and ETSFormer \citep{woo2022etsformer}, a transformer that replaces self-attention with exponential smoothing attention and frequency attention. We finally benchmark against OneFitsAll \citep{zhou2023one}, a method that leverages a pretrained large language model (LLM) (GPT-2 \citep{radford2019language}) and finetunes the input and output layers for time series forecasting.

Note that all the methods are compared in the \emph{univariate} setup, where, similar to \model, each time series is treated and forecasted independently. All methods produced using AutoGluon support probabilistic forecasts. All the other models (N-BEATS, Informer, AutoFormer, ETSFormer, and OneFitsAll) were originally designed for point forecasting and clean normalized data; we adapt them for probabilistic forecasting by using a distribution head at the output and endowing them with all the features similar to \model such as value scaling. 
    

%
    
\subsection{Hyperparameter Search and Model Training Setups}

We perform a random search of $100$ different hyperparameter configurations and use the validation loss of the pretraining corpus to select our model. We elaborate on our hyperparameter search and model selection in Appendix \ref{app:lag-llama-hyperparameters}. During \textbf{pretraining}, we use the batch size of $256$ and a learning rate of $10^{-4}$. Each epoch consists of $100$ randomly sampled windows, each of length $L+C$ as described in Sec. \ref{sec:lagFeatures}. We use an early stopping criterion of $50$ epochs based on the average validation loss of the training datasets in our pretraining corpus. When \textbf{fine-tuning} for a specific dataset, we train our models with the same batch size and learning rate, and each epoch consists of $100$ randomly sampled windows from the specific dataset, each of length $L + (C + P)$, where $P$ now is the prediction length of the specific dataset. Since our model is decoder-only, and since  prediction length is not fixed, the  model can  therefore work for any  downstream prediction length. We use an early stopping criterion of $50$ epochs during fine-tuning, based on the validation loss of the dataset being finetuned on. We elaborate on our training procedure in Appendix \ref{app:protocol-details}. For all the models trained in this paper, we use a single Nvidia Tesla-P100 GPU with 12 GB of memory, 4 CPU cores, and 24 GB of RAM. 


\begin{table*}[t]

\caption{CRPS of \model on few-shot adaptation on the {unseen datasets} with different amounts of data history being available, compared to supervised baselines trained solely on the respective datasets. Lower is better. A mean or standard deviation of 0.0000 signifies that the first non-zero digit is beyond $3$ decimal places. The best results are in \textbf{bold}.} 
\label{table:performance-fewshot}
\vskip 0.15in
\begin{center}
\begin{scriptsize}
\begin{sc}
\addtolength{\tabcolsep}{-0.6em}
\begin{tabular}{lccccccccc}
\toprule
\multirow{2}{*}{\textbf{Data \%}} & \multirow{2}{*}{\textbf{Model}} & \multicolumn{7}{c}{\textbf{Dataset}} & \multirow{2}{*}{\textbf{Average Rank}} \\
\cmidrule(lr){3-9}
& & \texttt{weather} & \texttt{ped-counts} & \texttt{exchange-rate} & \texttt{ett-m2} & \texttt{platform-delay} & \texttt{requests} & \texttt{beijing-pm2.5} & \\

\midrule
\multirow{4}{*}{{20 \%}} & DeepAR & 0.156$\pm$0.004 & 0.241$\pm$0.002 & 0.033$\pm$0.000 & 0.089$\pm$0.000 & 0.094$\pm$0.002 & 0.065$\pm$0.000 & 0.176$\pm$0.006 & 3.429 \\
& PatchTST & 0.169$\pm$0.017 & 0.259$\pm$0.008 & 0.012$\pm$0.000 & 0.035$\pm$0.001 & 0.088$\pm$0.001 & 0.025$\pm$0.000 & 0.153$\pm$0.003 & 2.714 \\
& TFT & 0.154$\pm$0.002 & 0.296$\pm$0.027 & \textbf{0.009$\pm$0.000} & 0.038$\pm$0.000 & \textbf{0.087$\pm$0.002} & {0.017$\pm$0.000} & \textbf{0.144$\pm$0.004 }& 2.000 \\
& Lag-Llama & \textbf{0.136$\pm$0.001} & \textbf{0.239$\pm$0.016} & 0.017$\pm$0.001 & \textbf{0.016$\pm$0.001} & 0.108$\pm$0.005 & \textbf{0.011$\pm$0.001} & 0.147$\pm$0.008 & 1.857 \\

\midrule

\multirow{4}{*}{{40 \%}} & DeepAR & 0.159$\pm$0.022 & 0.237$\pm$0.022 & 0.011$\pm$0.002 & 0.053$\pm$0.000 & 0.100$\pm$0.000 & 0.030$\pm$0.003 & 0.158$\pm$0.000 & 3.071 \\
& PatchTST & 0.171$\pm$0.017 & 0.253$\pm$0.007 & 0.011$\pm$0.001 & 0.035$\pm$0.000 & \textbf{0.092$\pm$0.000} & {0.025$\pm$0.002} & 0.162$\pm$0.000 & 2.929 \\
& TFT & 0.156$\pm$0.001 & 0.269$\pm$0.002 & \textbf{0.008$\pm$0.000} & 0.036$\pm$0.000 & 0.104$\pm$0.000 & 0.014$\pm$0.002 & 0.150$\pm$0.000 & 2.500 \\
& Lag-Llama & \textbf{0.135$\pm$0.000} & \textbf{0.229$\pm$0.003} & 0.009$\pm$0.001 & \textbf{0.017$\pm$0.002} & 0.102$\pm$0.002 &\textbf{ 0.014$\pm$0.001 }& \textbf{0.149$\pm$0.011} & 1.500 \\
        
\midrule

\multirow{4}{*}{{60 \%}} & DeepAR & {0.158$\pm$0.023} & \textbf{0.234$\pm$0.009 }& 0.011$\pm$0.001 & 0.049$\pm$0.006 & 0.114$\pm$0.006 & 0.026$\pm$0.002 & 0.157$\pm$0.004 & 3.071 \\
 & PatchTST & 0.174$\pm$0.011 & 0.241$\pm$0.004 & 0.011$\pm$0.000 & 0.035$\pm$0.001 & \textbf{0.093$\pm$0.003} & {0.028$\pm$0.002} & 0.159$\pm$0.001 & 2.929  \\
 & TFT & 0.152$\pm$0.001 & 0.272$\pm$0.000 & \textbf{0.008$\pm$0.000} & 0.037$\pm$0.000 & 0.113$\pm$0.008 & 0.017$\pm$0.002 & 0.154$\pm$0.000 & 2.429  \\
& Lag-Llama & \textbf{0.133$\pm$0.001 }& 0.246$\pm$0.002 & 0.009$\pm$0.001 & \textbf{0.016$\pm$0.001} & 0.099$\pm$0.005 & \textbf{0.012$\pm$0.001 }& \textbf{0.133$\pm$0.003} & 1.571  \\

\midrule

\multirow{4}{*}{{80 \%}} & DeepAR & 0.145$\pm$0.005 & 0.243$\pm$0.015 & 0.016$\pm$0.003 & 0.071$\pm$0.020 & 0.113$\pm$0.002 & 0.131$\pm$0.000 & 0.156$\pm$0.001 & 3.429 \\
& PatchTST & 0.174$\pm$0.033 & 0.247$\pm$0.015 & 0.015$\pm$0.002 & 0.035$\pm$0.000 & 0.091$\pm$0.003 & {0.024$\pm$0.000} & 0.153$\pm$0.002 & 2.714 \\
& TFT & 0.148$\pm$0.004 & 0.287$\pm$0.013 & \textbf{0.008$\pm$0.000} & 0.042$\pm$0.008 & \textbf{0.094$\pm$0.001} & 0.017$\pm$0.000 & 0.152$\pm$0.006 & 2.429 \\
& Lag-Llama & \textbf{0.132$\pm$0.001} & \textbf{0.215$\pm$0.006} & 0.009$\pm$0.000 & \textbf{0.019$\pm$0.001} & 0.099$\pm$0.008 & \textbf{0.013$\pm$0.002} & \textbf{0.131$\pm$0.016} & 1.429 \\

\bottomrule
\end{tabular}
\end{sc}
\end{scriptsize}
\end{center}
\vskip -0.1in
\end{table*}

\subsection{Inference and Model Evaluation}


Inference for a specific dataset is performed by sampling from the \model model autoregressively, starting with conditioning on the context of length $C$, until a prediction length $P$, which is defined for a given dataset. We use the Continuous Ranked Probability Score (CRPS) \citep{doi:10.1198/016214506000001437, RePEc}, a common metric in the probabilistic forecasting literature \citep{rasul2021multivariate, pmlr-v139-rasul21a, SalinasGPVar, shchur2023autogluontimeseries}, for evaluating our model's performance. We use 100 empirical samples and report the CRPS averaged over the prediction horizon and across all the time series of a dataset. We further assess how well each method we benchmark on does as a \textit{general-purpose forecasting algorithm}, rather than a \textit{dataset-specific one}, by measuring the average rank of each method, with respect to all others, over all the datasets.

\section{Results}

We first evaluate \textit{zero-shot} performance of our pretrained \model  on the unseen datasets  (subsection \ref{subsec:zeroshot-performance}), when no samples from the new downstream domain are available for possible fine-tuning  of the the model.
Note that such zero-shot forecasting scenarios are common in time series forecasting literature (see,  for example, the cold-start problem \citep{wiki:ColdStart, Fatemi2023MitigatingCF}). We then \textit{fine-tune} our pretrained \model on each unseen dataset and evaluate the model after fine-tuning, to study how our pretrained model adapts to different unseen datasets and domains when there is considerable history available in the dataset to train on. We then evaluate the \textit{few-shot} adaptation performance of our foundation model --- a well-known scenario in other modalities (e.g., text) where foundation models are expected to demonstrate strong generalization capabilities. We vary the amount of history available for fine-tuning on each dataset, and present the few-shot adaptation performance of our model at various levels of history (section \ref{subsec:fewshot-performance}). 


\subsection{Zero-Shot \& Finetuning Performance on New Data} \label{subsec:zeroshot-performance} 

\cref{table:performance} presents the results comparing the performance of supervised baselines trained on specific datasets to   the pretrained \model \textit{zero-shot} performance on the unseen datasets, and to finetuned \model on the respective unseen datasets.
In the \textit{zero-shot} setting, \model achieves comparable performance to all baselines, with an average rank of $6.714$. On \textit{fine-tuning}, \model achieves {state-of-the-art} performance in $3$ datasets, while performance increases significantly in all other datasets. Most importantly, on fine-tuning, \model achieves the \textbf{best average rank} of $2.786$, with a significant difference of $2$ points over the best supervised model, which suggests that if one had to choose a method to use without prior knowledge of the data, \model would be the best option. This clearly establishes \model as a strong foundation model that can be used on a wide range of downstream datasets, without prior knowledge of these data distribution --- a key property that a foundation model should satisfy.

We now take a deeper dive into  \model's performance analysis. Evaluated \textit{zero-shot}, \model achieves strong performance, notably in the platform-delay and weather datasets, where it is especially close to baselines. With fine-tuning, \model consistently improves performance compared to inferring zero-shot. In $3$ datasets - namely, ETT-M2, weather, and requests --- finetuned version of  \model achieves a significantly lower error than all the baselines, becoming the  {state-of-the-art}. On the \textit{exchange-rate} dataset coming from an entirely new domain, exhibiting a new unseen frequency, \model has comparable zero-shot performance, and when finetuned achieves performance similar to the state-of-the-art. This establishes that \model performs well across frequencies and domains from which the model may or may not have seen similar data on during pretraining.
\model achieves a better average rank both in the \textit{zero-shot} and \textit{finetuned} setups compared to the Informer, AutoFormer, and ETSFormer models, all of which use complex inductive biases to model time series,  compared to \model which uses a simple architecture, lags and covariates, along with large-scale pretraining. Our observations suggest that \textbf{at scale}, when used similarly to \model, vanilla decoder-only transformers outperform other transformer architectures. We point out that similar results have been shown in the NLP community \citep{tay2022scaling} studying the influence of inductive bias at scale, however, we emphasize that {we are the first to point out} such a result for time series, potentially opening doors to further studies in time series that analyse the influence of inductive bias at scale. Next, compared to the OneFitsAll model \citep{zhou2023one} which adapts a pretrained LLM for forecasting, \model achieves significantly better performance in all datasets, except for the dataset  beijing-pm2.5,  where it performs similarly to the baseline, while achieving a much better average rank than this model. These results demonstrate the potential of foundation models  trained \textit{from scratch} on a large and diverse collection of time series datasets when compared to the adaptation of pretrained  LLMs, as in the OneFitsAll model \citep{zhou2023one}. A detailed investigation of the advantages and disadvantages of adapting LLMs versus training time series foundation models from scratch is left as a direction for future work. 

We further visualize the forecasts produced by \model on the unseen datasets qualitatively in App. \cref{app:forecast-visualizations}. \model produces forecasts that closely match the ground truth. Further, comparing the forecasts produced by the model in the zero-shot (\cref{fig:Zeroshot-forecast-requests}) and fine-tuned (\cref{fig:FT-forecast-requests}) settings, one can clearly see that the quality of forecasts increase significantly when the model is fine-tuned. 

\subsection{Few-Shot Adaptation Performance on Unseen Data} \label{subsec:fewshot-performance} 

We restrict the data to only the last $K\%$ of the history from the training set of the datasets, where we set $K$ to $20$, $40$, $60$, $80$ percentages respectively. We train the supervised methods from scratch on the available data, while we fine-tune \model. Results are presented in \cref{table:performance-fewshot}.
Across varying levels of history being available for adaptation, \model achieves the {best average rank} across all levels, which establishes \model as one with strong adaptation capabilities across all levels of data. As the amount of history available increases, \model achieves increasingly better performance across all datasets, and the gap between the rank of \model and the baselines widens, as expected. 
Note, however, that \model is most often outperformed by TFT in the exchange-rate dataset, which is from an entirely new domain and has a new unseen frequency. Our observation demonstrates that, in cases where the data is most dissimilar, as compared to the pretraining corpus, \model requires increasing amounts of history to train on, and, when given enough history to adapt, performs comparable to {state-of-the-art} (as discussed in subsection \ref{subsec:zeroshot-performance}). 

\textit{Overall, our empirical results demonstrate that \model has strong few-shot adaptation capabilities, and that, based on the characteristics of the downstream dataset, \model can adapt and generalize with the appropriate amount of data.}

\section{Analysis} 
\subsection{Data Diversity} \label{sec:dataset-analysis}

Although loss has been found to scale with pre-training dataset size \cite{kaplan2020scaling}, it remains unclear what other properties of pre-training datasets lead to desirable model behaviour, despite some initial research in this direction \cite{chan2022data}.
Notably, diversity in the pre-training data has contributed to improved zero-shot performance and few-shot adaptation \cite{brown2020language}, notwithstanding the absence of an adequate definition.

To quantify the diversity of the pretraining corpus, we analyze the properties of its datasets through 22 Canonical time series Characteristics (``catch22 features''), a set of quickly computable time series features selected for their classification ability \cite{lubba2019catch22} from the features of the Highly Comparable Time Series Analysis (hctsa) library \cite{fulcher2013highly}.
To assess diversity across datasets, we apply PCA to the features averaged per-dataset and plot the top 2 components . 
We find that having multiple datasets within domains and across domains increases the diversity of AC22 features in the top 2-component space (see Figure \ref{fig:pca} in Appendix).

\subsection{Scaling Analysis}

Dataset size has been shown empirically to improve performance \cite{kaplan2020scaling}. Constructing neural scaling laws \cite{kaplan2020scaling, caballero2023broken} can help understand how the performance of the model scales with respect to different parameters such as the amount of pretraining data, number of parameters in the model etc.
Towards understanding these quantities for models such as \model, we fit neural scaling laws \cite{caballero2023broken} to our model's validation loss and present in App. \cref{app:neural-scaling-laws-contd} the obtained scaling laws that describe the performance of our model with respect the amount of pretraining data.

\section{Discussion}
We present \model, a foundation model for univariate probabilistic time series forecasting based on a simple decoder-only transformer architecture. We show that \model, when pretrained from scratch on a large corpus of datasets, has strong \textit{zero-shot} generalization performance on unseen datasets, and performs comparably to dataset-specific models. \model also demonstrates \textit{state-of-the-art} performance across diverse datasets from different domains after \textit{finetuning}, and emerges as the best general-purpose model without any knowledge of downstream datasets. \model also demonstrates a strong \textit{few-shot adaptation} performance across varying amounts  of data history being available. Finally, we investigate the diversity of the pretraining corpus used to train \model. 

Our work opens up several potential directions for future work. For now, collecting and collating a large scale time series corpus of \emph{open} dataset would be of high value, since the largest time series dataset repositories \citep{godahewa2021monash} are themselves too small. Further, scaling up the models further beyond the model sizes explored in this work using different training strategies constitutes an essential next step towards building even more powerful time series foundation models. Finally, expanding our work from univariate towards multivariate approaches by capturing complex multivariate dynamics of real-world datasets also constitutes an important direction for future work.


\section{Impact Statement}
The goal of this work is to introduce general-purpose foundation models for time series forecasting. There are many potential societal consequences of such models, including positive impacts on optimizing processes via better decision-making, as well as possible negative impacts. 

To the best of our knowledge, none of the datasets used contain nor are linked to any individual or personally identifiable data, and have been sourced from referenced locations.

\section{Contributions}

\textbf{Arjun} organized, planned, and led the project overall; refined and improved the \model architecture by refining key components (lags, sampling of the model), and refining the architecture and training strategies (such as dropout, early stopping, learning rate scheduling), iterated on the dataset choices for \model and dataset splitting strategies, fixed issues with data window sampling, ran and iterated on all large-scale pretraining, fine-tuning and few-shot learning experiments for \model, and wrote several main parts of the paper.

\textbf{Kashif} wrote the code for the \model architecture and training strategies, conducted initial experiments for \model and other lag-based architectures that were explored in the project; added Monash time series repository dataset to Hugging Face datasets as well as other datasets; implemented all (but one) transformer-based time series models; worked to merge fixes/features upstream to GluonTS; integrated code with Hugging Face for open-source release; and wrote several main parts of the paper.

\textbf{Hena} added support for time features, updated the alternative Lag-Transformer model for experiments, added support for the key-value cache for faster inference, compiled a list of all GluonTS datasets and their descriptions, and contributed to dataset compilation efforts, added utilities to track per-dataset validation and training loss, worked with the Informer, Autoformer and ETSFormer models for the paper for the large-scale experiments of the paper.

\textbf{Andrew} expanded the empirical design of the paper for the fine-tuning and downstream adaptation settings, ran experiments and contributed to the writing of the first version of the paper, wrote several key sections of the paper, adapted air quality and Huawei datasets, integrated robust scaler for data normalization, worked on the ideation and codebase of the Catch-22 feature-based dataset analysis for the paper.

\textbf{Rishika} ran experiments, and contributed to the writing of the first version of the paper, added all time series augmentations to the codebase of the paper, updated the alternative Lag-Transformer model for new experiments, adapted ETT datasets, Azure/Borg/Alibaba datasets (Cloud datasets), added options for automatic batch size search and plotting forecasts, integrated distribution heads such as IQN for experiments.

\textbf{Arian} ran experiments and contributed to the writing of the first version of the paper, worked with the OneFitsAll model initial code and experiments, and worked with the experiments for the N-BEATS model.

\textbf{Mohammad} worked with all AutoGluon models and experiments, added the option to use Stochastic Weight Averaging (SWA), and brainstormed about early stopping techniques to use when pretraining.

\textbf{George} ran experiments, and contributed to the writing of the first version of the paper, adapting the Electricity Household Consumption Dataset, M5, Walmart, Rossman, and Corporation and Restaurant Datasets used in the experiments of the project and the paper.
 
\textbf{Roland} worked with the code and experiments of the OneFitsAll model for all large-scale experiments in the paper, and contributed to writing several sections of the paper.
 
\textbf{Nadhir} integrated the N-BEATS model and worked with it for all large-scale experiments in the paper.

\textbf{Marin} wrote the initial code for sampling windows for the pretraining set and provided feedback with GluonTS code and experimental setups.

\textbf{Sahil, Anderson, Nicolas, Alexandre, Valentina, and Yuriy} advised the project as a whole, provided feedback on the experiments and the paper, and contributed to the writing of several sections of the paper.

\textbf{Irina} advised the project with feedback in several stages, contributing to the writing of the paper, acquisition of the funding for the project, and conceiving and pushing forward the research direction in the early stages of the project.




\section{Acknowledgements}

We are grateful to
Viatcheslav Gurev, 
for useful discussions during the course of the project.
We acknowledge and thank the authors and contributors of all the open-source libraries that were used in this work, especially: GluonTS \citep{gluonts_jmlr}, NumPy \citep{harris2020array}, Pandas \citep{reback2020pandas}, Matplotlib \citep{Hunter:2007} and  PyTorch \citep{NIPS2019_9015}.

We acknowledge the support from the Canada CIFAR AI Chair Program and from the Canada Excellence Research Chairs (CERC)
Program. This project used compute resources provided by the Oak Ridge Leadership Computing Facility at the Oak Ridge National Laboratory, which is supported by the Office of Science of the U.S. Department of Energy under Contract No. DE-AC05-00OR22725. This project further used compute resources provided by ServiceNow, Mila, and Compute Canada.

\bibliographystyle{icml2024}
\bibliography{references}

\clearpage

\newpage

\appendix


\section{Details of Datasets} \label{app:dataset-details}

\begin{table*}[]
\caption{Datasets used in the pretraining corpus and the unseen datasets on which we evaluate, grouped by the domains they are labelled against.}
\label{table:datasets-domain}
\resizebox{\textwidth}{!}{%
\begin{tabular}{@{}lllllll@{}}
\toprule
 & \multicolumn{1}{c}{Transport \& Tourism} & \multicolumn{1}{c}{Energy}  & \multicolumn{1}{c}{Nature} & \multicolumn{1}{c}{Air Quality} & \multicolumn{1}{c}{Cloud} & \multicolumn{1}{c}{Banking \& Econ} \\ \midrule
Pretraining & San Francisco Traffic & Australian Electricity Demand  & KDD Cup 2018 & Beijing Multisite & CPU Limit Minute \\ 
 & Uber TLC Hourly & Electricity Hourly & Sunspot & UCI & CPU Usage Minute \\
 &  & London Smart Meters  &  &  & Function Delay Minute \\
 &  & Solar &  &  &  Instances Minute  \\
 &  & Wind Farms &  &   & Memory Limit Minute \\
 &  & ETT H1 &  &  &  Memory Usage Minute \\
 &  & ETT H2 &  &  &   \\ 
 &  & ETT M1 &  &  &   \\ \bottomrule
Unseen & Pedestrian Counts & ETT M2 & Weather & Beijing PM2.5 & Requests Minute & Exchange Rate  \\ 
 &  &  &  &  &  Platform Delay Minute \\ \midrule
\end{tabular}%
}
\end{table*}

\begin{table*}[]
\caption{Statistics of all the datasets used in the paper. Frequencies \texttt{H} stands for Hourly, \texttt{T} for minute, and \texttt{B} for business day. Tokens refers to the total number of windows of size $1$ in the dataset, computed as the aggregate number of timesteps across all series in that dataset.}
\resizebox{\textwidth}{!}{%
\begin{tabular}{@{}llccrrr@{}}
\toprule
\multicolumn{1}{c}{\multirow{2}{*}{Dataset}} &
  \multicolumn{1}{c}{\multirow{2}{*}{Freq}} &
  \multicolumn{1}{c}{\multirow{2}{*}{Domain}} &
  \multicolumn{1}{c}{\multirow{2}{*}{Prediction Length}} &
  \multicolumn{3}{c}{Train split} \\ \cmidrule(l){5-7} 
\multicolumn{1}{c}{} &
  \multicolumn{1}{c}{} &
  \multicolumn{1}{c}{} &
  \multicolumn{1}{c}{} &
  \multicolumn{1}{c}{Timestamps} &
  \multicolumn{1}{c}{\# Series} &
  \multicolumn{1}{c}{Tokens} \\ \midrule
Australian Electricity Demand & 0.5H & Energy      & 60 & 230676 & 5    & 1153380   \\
Electricity Hourly            & H    & Energy      & 48 & 26256  & 321  & 8428176   \\
London Smart Meters           & 0.5H & Energy      & 60 & 23844  & 5560 & 132572640 \\
Solar                         & 10T  & Energy      & 60 & 52500  & 137  & 7192500   \\
Wind Farms                    & T    & Energy      & 60 & 526980 & 339  & 178646220 \\
Pedestrian Counts             & H    & Transport   & 48 & 84283  & 66   & 5562678   \\
Uber TLC Hourly               & H    & Transport   & 24 & 4254   & 262  & 1114548   \\
Traffic                       & H    & Transport   & 24 & 14036  & 862  & 12099032  \\
KDD Cup 2018                  & H    & Nature      & 48 & 10850  & 270  & 2929500   \\
Sunspot                       & D    & Nature      & 30 & 73894  & 1    & 73894     \\
Weather                       & D    & Nature      & 30 & 695    & 3010 & 2091950   \\
Exchange Rate                 & 1B   & Economic    & 30 & 6071   & 8    & 48568     \\
ETT H1                        & H    & Energy      & 24 & 8640   & 1    & 8640      \\
ETT H2                        & H    & Energy      & 24 & 8640   & 1    & 8640      \\
ETT M1                        & 15T  & Energy      & 24 & 34560  & 1    & 34560     \\
ETT M2                        & 15T  & Energy      & 24 & 34560  & 1    & 34560     \\
Requests Minute               & T    & Cloud       & 60 & 64800  & 10   & 648000    \\
Function Delay Minute         & T    & Cloud       & 60 & 64800  & 10   & 648000    \\
Platform Delay Minute         & T    & Cloud       & 60 & 64800  & 10   & 648000    \\
CPU Usage Minute              & T    & Cloud       & 60 & 64800  & 10   & 648000    \\
Memory Usage Minute           & T    & Cloud       & 60 & 64800  & 10   & 648000    \\
CPU Limit Minute              & T    & Cloud       & 60 & 64800  & 10   & 648000    \\
Memory Limit Minute           & T    & Cloud       & 60 & 64800  & 10   & 648000    \\
Instances Minute              & T    & Cloud       & 60 & 64800  & 10   & 648000    \\
UCI                           & H    & Air Quality & 24 & 9357   & 13   & 121641    \\
Beijing PM2.5                  & H    & Air Quality & 24 & 43824  & 8    & 350592    \\
Beijing Multisite             & H    & Air Quality & 24 & 35064  & 132  & 4628448   \\ \bottomrule
                        
\end{tabular}%
}
\label{tab:dataset-table}
\end{table*}

\begin{table*}[!htp]
\caption{Hyperparameter choices for \model. The values with * represent the optimal values obtained by hyperparameter search. \\$^\dag$ Note that this is just the consecutive context that is sampled for each window; in practice we use a much larger context window due to the use of lags, as described in Sec. \cref{sec:lagFeatures}  }
\label{tab:lag-llama-hps}
\vskip 0.15in
\begin{center}
\begin{small}
\begin{sc}
\begin{tabular}{@{}lc}
\toprule
\textbf{Hyperparameter}      & \multicolumn{1}{c}{\model}       \\
\midrule                                                                                  
Number of layers               & 1,2,3,4,5,6,7,8*,9     \\
Number of heads             & 1,2,3,4,5,6,7,8,9*    \\
Embedding Dimensions per head         & 16*, 32, 64, 128, 256, 512   \\
Context Length $C$ $^\dag$  & 32*, 64, 128, 256, 512, 1024  \\
Augmentation Probability          & 0,0.25,0.5*,1.0    \\
Frequency Masking Rate          & 0,0.25,0.5*,1.0   \\
Frequency Mixing Rate          & 0,0.25*,0.5,1.0 \\
Weight Decay      & 0*,0.25,0.5,1.0    \\
Dropout          & 0*,0.25,0.5,1.0   \\
\bottomrule
\end{tabular}
\end{sc}
\end{small}
\end{center}
\vskip -0.1in
\end{table*}

We use the following datasets in our experiments, the statistics of which are in Table \ref{tab:dataset-table}, and their domains in Table \ref{table:datasets-domain}. Table \ref{table:datasets-domain} further presents if a dataset was present in the pretraining or downstream testing corpora in our work.

The \textbf{Air Quality UC Irvine Repository} dataset (UCI) contains 9358 instances of hourly averaged responses from 5 metal oxide chemical sensors embedded in an Air Quality Chemical Multisensor Device in a polluted area \cite{misc_air_quality_360}. 

The \textbf{Australian Electricity Demand} dataset comprises five half-hourly time series of the electricity demand across five Australian states: Victoria, New South Wales, Queensland, Tasmania, and South Australia \cite{godahewa2021monash}.

The \textbf{Beijing PM2.5} dataset contains hourly data of PM2.5 levels recorded by the US Embassy in Beijing. 
The dataset also includes meteorological data from Beijing Capital International Airport \cite{misc_beijing_pm2.5_data_381}. 

The \textbf{Beijing Multi-Site Air-Quality} dataset comprises hourly measurements of six primary air pollutants and six corresponding meteorological variables at various locations in Beijing over a period of four years. \cite{misc_beijing_multi-site_air-quality_data_501}

The \textbf{Electricity Hourly} dataset captures electricity usage for 321 clients measured at hourly intervals from 2012 to 2014 \cite{godahewa2021monash}.

The \textbf{ETTh1, ETTh2, ETTm1, ETTm2} datasets contain 2 years worth of data obtained from two Electricity Transformers at hourly and 15-minute frequencies curated to help predict if electrical transformers’ oil is at a safe temperature \cite{ett-haoyietal-informer-2021}.

The \textbf{Exchange Rate} compilation encompasses the daily exchange rates of eight foreign currencies, namely Australia, the United Kingdom, Canada, Switzerland, China, Japan, New Zealand, and Singapore, spanning the period from 1990 to 2016 \cite{godahewa2021monash}.

The \textbf{Huawei cloud} datasets contain serverless traces \cite{cloud-datasets}. 
We select $8$ series containing metrics based on the minute-frequency occurrences of the top 10 functions by median occurrences over 141 days:
\textbf{function delay,
platform delay,
cpu usage,
memory usage,
cpu limit,
memory limit,
instances.
platform delay,
requests}.

The  \textbf{London Smart Meters} dataset focuses on electrical consumption readings from smart meters in 5,567 households that participated in the UK Power Networks Low Carbon London project between November 2011 and February 2014 \cite{godahewa2021monash}. 

The \textbf{KDD Cup 2018} dataset comprises extensive hourly time series data reflecting air quality levels across 59 stations in Beijing and London from January 2017 to March 2018. 
Measurements include PM2.5, PM10, NO2, CO, O3, and SO2 \cite{godahewa2021monash}. 

The \textbf{Pedestrian Counts} dataset (referred to as ped-counts in parts of the text) encompasses hourly pedestrian counts recorded by 66 sensors within the city of Melbourne, commencing in May 2009 \cite{godahewa2021monash}.

The \textbf{Solar} dataset comprises 6000 simulated time series for 5-minute solar power and hourly forecasts of photovoltaic  power plants in the U.S. in 2006. 
It includes 137 time series reflecting solar power production every 10 minutes in Alabama during 2006 \cite{godahewa2021monash}.

The \textbf{Sunspot} dataset comprises a singular extensive daily time series of sunspot numbers spanning from January 1818 to May 2020 \cite{godahewa2021monash}.

The \textbf{Traffic} dataset encompasses 862 hourly time series depicting road occupancy rates on the freeways in the San Francisco Bay area from 2015 to 2016 \cite{godahewa2021monash}.

The \textbf{Uber TLC Hourly} dataset consists data of 4.5 million Uber pickups in NYC (April-September 2014) and 14.3 million pickups (January-June 2015). 
It includes trip details for 10 other for-hire vehicle companies and aggregated data for 329 companies \cite{538-uber-tlc, godahewa2021monash}.

The \textbf{Weather} dataset includes time series of hourly climate data near Monash University, Clayton, Victoria, Australia, from January 2010 to May 2021.
The data contains series for temperature, dewpoint temperature, wind speed, mean sea level pressure, relative humidity, surface solar radiation, surface thermal radiation, and total cloud cover \cite{godahewa2021monash}.

The \textbf{Wind Farms} dataset contains minute-frequency time series data tracking the wind power production of 339 wind farms in Australia \cite{godahewa2021monash}.


\section{Protocol Details} \label{app:protocol-details}


For all datasets used in the paper, we have a training and test split that are non-overlapping based on the timestamps, as defined in the dataset. \textbf{During pretraining}, for each such dataset, we exclude the $14$ last overlapping windows of the train split, and use it as the dataset's validation set. When pretraining, we train on a combined dataset formed out of the train split of each dataset, after every epoch, we obtain the validation loss on the validation sets of all datasets used in the pretraining corpus. We use the average validation loss for early stopping criterion (this is referred to as "validation loss" in the paper). \textbf{When fine-tuning on a specific dataset}, we exclude the single last window of the train split, and use it as the dataset's validation set. We train on the train split of the dataset, and use the validation split for early stopping. We use the same setup as fine-tuning \model, for all supervised baselines that we produce results for in the paper. Following typical evaluation setups \citep{shchur2023autogluontimeseries}, all results reported in the paper are on the last prediction window of the test splits defined in App. \cref{app:dataset-details}.



\section{Additional Empirical Results} \label{app:addnl-empirical-results}

\subsection{Results on the Pretraining Datasets}

A strong foundation model should not just be good at adapting zero-shot and few-shot to unseen distributions of data, but should also perform well \textit{in-distribution}, i.e. on the datasets that the model has been pretrained on. Therefore, apart from evaluating our model on unseen datasets, we also evaluate our model on those datasets we use for pretraining. 

Results are given in \cref{table:training-performance-1}, \cref{table:training-performance-2}, and \cref{table:training-performance-3}. Results on Average Rank on all datasets are given in \cref{table:training-performance-av-rank}. The training budget of \model was split among all the pretraining datasets, while other supervised models on the dataset do not have that constraint. Thereby, \model did not see as much data in each dataset as the other models, and thereby is not expected to perform as well as each supervised model on the specific datasets. This is reflected in the results, as \model is not the best performing model in each dataset. Still, \model achieves a comparable average rank, and is among the models achieving the top average ranks. 

\begin{table*}[t]

\caption{\textit{CRPS of \model on $7$/$20$ datasets in the pretraining corpus, compared to supervised baselines trained solely on the respective datasets. Lower is better. A mean or standard deviation of 0.0000 signifies that the first non-zero digit is beyond $3$ decimal places.}} 
\label{table:training-performance-1}
\vskip 0.15in
\begin{center}
\begin{scriptsize}
\begin{sc}
\addtolength{\tabcolsep}{-0.6em}
\begin{tabular}{lccccccc}
\toprule
\multirow{2}{*}{\textbf{Model}} & \multicolumn{7}{c}{\textbf{Dataset}} \\
\cmidrule(lr){2-8}
& \texttt{aus-elec-demand} & \texttt{electricity}  & \texttt{kdd-cup} & \texttt{london-smart-meters} & \texttt{solar} & \texttt{sunspot} & \texttt{traffic} \\

\midrule
AutoARIMA & 0.065$\pm$0.000 & 0.098$\pm$0.003 & 0.552$\pm$0.000 & nan$\pm$nan & 0.558$\pm$0.000 & 77.862$\pm$0.000 & 0.277$\pm$0.000\\
AutoETS & 0.160$\pm$0.000 & 0.104$\pm$0.000 & 2.350$\pm$0.000 & nan$\pm$nan & 0.551$\pm$0.000 & 171.363$\pm$0.000 & 0.492$\pm$0.000\\
CrostonSBA & 0.127$\pm$0.000 & 0.244$\pm$0.000 & 0.459$\pm$0.000 & 0.500$\pm$0.000 & 1.016$\pm$0.000 & 34.458$\pm$0.000 & 0.414$\pm$0.000\\
DeepAR & 0.043$\pm$0.000 & 0.085$\pm$0.005 & 0.327$\pm$0.014 & 0.409$\pm$0.000 & 0.446$\pm$0.002 &\textbf{ 1.390$\pm$0.000} & \textbf{0.100$\pm$0.000}\\
DynamicOptimize & 0.043$\pm$0.000 & 0.203$\pm$0.000 & 0.550$\pm$0.000 & 0.681$\pm$0.000 & 1.580$\pm$0.000 & 181.350$\pm$0.000 & 0.383$\pm$0.000\\
NPTS & 0.098$\pm$0.000 & 0.139$\pm$0.001 & 0.346$\pm$0.001 & 0.464$\pm$0.000 & \textbf{0.404$\pm$0.001} & 201.558$\pm$10.653 & 0.191$\pm$0.000\\
PatchTST & 0.056$\pm$0.000 & 0.088$\pm$0.001 & 0.432$\pm$0.043 & 0.375$\pm$0.000 & 0.734$\pm$0.002 & 3.083$\pm$0.000 & 0.153$\pm$0.001\\
TemporalFusionT & 0.041$\pm$0.000 & 0.100$\pm$0.008 & 0.411$\pm$0.023 & 0.343$\pm$0.000 & 0.443$\pm$0.003 & 25.675$\pm$0.000 & 0.108$\pm$0.001\\
NBEATS & \textbf{0.032$\pm$0.002} &\textbf{ 0.072$\pm$0.000} & 0.435$\pm$0.080 & 0.453$\pm$0.000 & 0.655$\pm$0.000 & 20.089$\pm$20.404 & 0.116$\pm$0.000\\
OFA & 0.112$\pm$0.003 & 0.286$\pm$0.040 & 0.491$\pm$0.034 & \textbf{0.285$\pm$0.046} & 3.786$\pm$0.234 & 38.119$\pm$1.536 & 0.446$\pm$0.009\\
Informer & 0.064$\pm$0.020 & 0.081$\pm$0.002 & 0.351$\pm$0.000 & 0.424$\pm$0.011 & 0.990$\pm$0.140 & 4.765$\pm$0.336 & 0.157$\pm$0.000\\
AutoFormer & 0.090$\pm$0.021 & 0.102$\pm$0.005 & 0.451$\pm$0.018 & 0.383$\pm$0.003 & 2.107$\pm$0.425 & 40.456$\pm$12.354 & 0.185$\pm$0.010\\
ETSFormer & 0.105$\pm$0.011 & 0.191$\pm$0.026 & 0.692$\pm$0.071 & 0.460$\pm$0.009 & 1.271$\pm$0.086 & 58.708$\pm$17.080 & 0.188$\pm$0.008\\
LagLLama & 0.087$\pm$0.018 & 0.095$\pm$0.013 & \textbf{0.323$\pm$0.004} & 0.381$\pm$0.003 & 1.536$\pm$0.237 & 4.961$\pm$1.912 & 0.119$\pm$0.001 \\

\bottomrule
\end{tabular}
\end{sc}
\end{scriptsize}
\end{center}
\vskip -0.1in
\end{table*}

\begin{table*}[t]

\caption{\textit{CRPS of \model on the $7$/$20$ datasets in the pretraining corpus, compared to supervised baselines trained solely on the respective datasets. Lower is better. A mean or standard deviation of 0.0000 signifies that the first non-zero digit is beyond $3$ decimal places.}} 
\label{table:training-performance-2}
\vskip 0.15in
\begin{center}
\begin{scriptsize}
\begin{sc}
\addtolength{\tabcolsep}{-0.6em}
\begin{tabular}{lccccccc}
\toprule
\multirow{2}{*}{\textbf{Model}} & \multicolumn{7}{c}{\textbf{Dataset}} \\
\cmidrule(lr){2-8}
 & \texttt{uber} & \texttt{windfarms}  & \texttt{ETT\_h1}  & \texttt{ETT\_h2}  & \texttt{ETT\_m1}  & \texttt{AirQualityUCI}  & \texttt{BeijingMultisite} \\

\midrule
AutoARIMA & 0.322$\pm$0.000 & 0.084$\pm$0.000 & 0.120$\pm$0.000 & 0.095$\pm$0.000 & nan$\pm$nan & 0.206$\pm$0.000 & 0.359$\pm$0.000\\
AutoETS & 0.461$\pm$0.000 & 0.096$\pm$0.000 & 0.117$\pm$0.000 & 0.105$\pm$0.000 & 0.073$\pm$0.000 & 0.220$\pm$0.000 & 0.472$\pm$0.000\\
CrostonSBA & 0.427$\pm$0.000 & 0.130$\pm$0.000 & 0.123$\pm$0.000 & 0.112$\pm$0.000 & 0.094$\pm$0.000 & 0.237$\pm$0.000 & 0.400$\pm$0.000\\
DeepAR & 0.170$\pm$0.003 & 0.070$\pm$0.000 & 0.105$\pm$0.002 & 0.082$\pm$0.010 & 0.074$\pm$0.007 & 0.195$\pm$0.006 & 0.282$\pm$0.032\\
DynamicOptimize & 0.433$\pm$0.000 & 0.060$\pm$0.000 & 0.117$\pm$0.000 & 0.085$\pm$0.000 & 0.070$\pm$0.000 & 0.216$\pm$0.000 & 0.394$\pm$0.000\\
NPTS & 0.191$\pm$0.000 & 0.208$\pm$0.000 & 0.268$\pm$0.001 & 0.216$\pm$0.001 & 0.162$\pm$0.000 & \textbf{0.130$\pm$0.001} & 0.414$\pm$0.006\\
PatchTST & 0.219$\pm$0.007 & 0.057$\pm$0.000 & 0.099$\pm$0.001 & 0.067$\pm$0.001 & 0.063$\pm$0.001 & 0.189$\pm$0.003 & 0.304$\pm$0.016\\
TemporalFusionT & 0.197$\pm$0.012 & \textbf{0.055$\pm$0.000} & 0.082$\pm$0.006 & 0.049$\pm$0.001 & 0.058$\pm$0.000 & 0.227$\pm$0.026 & 0.410$\pm$0.019\\
NBEATS & 0.352$\pm$0.000 & 0.117$\pm$0.000 & \textbf{0.013$\pm$0.001} & \textbf{0.010$\pm$0.001} & \textbf{0.009$\pm$0.000 }& 0.156$\pm$0.004 & 0.340$\pm$0.016\\
OFA & 0.424$\pm$0.006 & 0.190$\pm$0.010 & 0.172$\pm$0.002 & 0.148$\pm$0.002 & 0.146$\pm$0.006 & 0.201$\pm$0.016 & 0.362$\pm$0.040\\
Informer & 0.196$\pm$0.003 & 0.099$\pm$0.014 & 0.174$\pm$0.003 & 0.112$\pm$0.014 & 0.098$\pm$0.008 & 0.191$\pm$0.024 & 0.241$\pm$0.016\\
AutoFormer & 0.205$\pm$0.007 & 0.246$\pm$0.038 & 0.155$\pm$0.010 & 0.119$\pm$0.005 & 0.119$\pm$0.008 & 0.172$\pm$0.012 & \textbf{0.238$\pm$0.012}\\
ETSFormer & 0.313$\pm$0.011 & 0.588$\pm$0.331 & 0.142$\pm$0.004 & 0.102$\pm$0.005 & 0.108$\pm$0.003 & 0.197$\pm$0.021 & 0.481$\pm$0.084\\
LagLLama &\textbf{ 0.168$\pm$0.002} & 0.145$\pm$0.009 & 0.104$\pm$0.001 & 0.073$\pm$0.005 & 0.068$\pm$0.001 & 0.138$\pm$0.006 & 0.340$\pm$0.055 \\

\bottomrule
\end{tabular}
\end{sc}
\end{scriptsize}
\end{center}
\vskip -0.1in
\end{table*}

\begin{table*}[t]

\caption{\textit{CRPS of \model on the $6$/$20$ datasets in the pretraining corpus, compared to supervised baselines trained solely on the respective datasets. Lower is better. A mean or standard deviation of 0.0000 signifies that the first non-zero digit is beyond $4$ decimal places.}} 
\label{table:training-performance-3}
\vskip 0.15in
\begin{center}
\begin{scriptsize}
\begin{sc}
\addtolength{\tabcolsep}{-0.6em}
\begin{tabular}{lcccccc}
\toprule
\multirow{2}{*}{\textbf{Model}} & \multicolumn{6}{c}{\textbf{Dataset}} \\
\cmidrule(lr){2-7}
 & \texttt{cpu\_limit} & \texttt{cpu\_usage} & \texttt{function\_delay} & \texttt{instances} & \texttt{memory\_limit} & \texttt{memory\_usage}\\
\midrule

AutoARIMA & 0.2245$\pm$0.0000 & 0.0814$\pm$0.0000 & 0.0936$\pm$0.0000 & 0.0121$\pm$0.0000 & 0.2024$\pm$0.0000 & 0.0326$\pm$0.0000\\
AutoETS & 0.0632$\pm$0.0000 & 0.0806$\pm$0.0000 & nan$\pm$nan & 0.0128$\pm$0.0000 & 0.0632$\pm$0.0000 & 0.0664$\pm$0.0000\\
CrostonSBA & 0.0278$\pm$0.0000 & 0.0826$\pm$0.0000 & 0.0756$\pm$0.0000 & 0.0318$\pm$0.0000 & 0.0278$\pm$0.0000 & 0.0346$\pm$0.0000\\
DeepAR & 0.0004$\pm$0.0001 & 0.1034$\pm$0.0016 & 0.1097$\pm$0.0039 & 0.0179$\pm$0.0054 & 0.0004$\pm$0.0000 & 0.0147$\pm$0.0016\\
DynamicOptimize & 0.0012$\pm$0.0000 & 0.0813$\pm$0.0000 &\textbf{ 0.0381$\pm$0.0000} & 0.0140$\pm$0.0000 & 0.0010$\pm$0.0000 & 0.0667$\pm$0.0000\\
NPTS & 0.0001$\pm$0.0001 & 0.1010$\pm$0.0004 & 0.0808$\pm$0.0008 & 0.0158$\pm$0.0002 & 0.0001$\pm$0.0001 & 0.0164$\pm$0.0002\\
PatchTST & 0.0023$\pm$0.0005 & \textbf{0.0805$\pm$0.0026} & 0.0571$\pm$0.0000 & 0.0104$\pm$0.0021 & 0.0042$\pm$0.0012 & 0.0172$\pm$0.0042\\
TemporalFusionT & \textbf{0.0001$\pm$0.0001} & 0.0830$\pm$0.0062 & 0.0552$\pm$0.0030 & \textbf{0.0057$\pm$0.0010 }& \underline{0.0000$\pm$0.0000} & 0.0113$\pm$0.0013\\
NBEATS & 0.0001$\pm$0.0000 & 0.0972$\pm$0.0018 & 0.0502$\pm$0.0030 & 0.0086$\pm$0.0012 & \underline{0.0000$\pm$0.0000} & 0.0121$\pm$0.0009\\
OFA & 0.0004$\pm$0.0003 & 0.1209$\pm$0.0082 & 0.1249$\pm$0.0170 & 0.0235$\pm$0.0019 & \underline{0.0000$\pm$0.0000} & 0.0137$\pm$0.0012\\
Informer & 0.0001$\pm$0.0000 & 0.0986$\pm$0.0040 & 0.0843$\pm$0.0143 & 0.0164$\pm$0.0000 & \underline{0.0000$\pm$0.0000} & 0.0110$\pm$0.0004\\
AutoFormer & 0.0392$\pm$0.0040 & 0.1040$\pm$0.0031 & 0.1652$\pm$0.0328 & 0.1311$\pm$0.0600 & 0.1489$\pm$0.0334 & \textbf{0.1301$\pm$0.1235}\\
ETSFormer & 0.0021$\pm$0.0015 & 0.1295$\pm$0.0061 & 0.2066$\pm$0.0774 & 0.5406$\pm$0.4268 & 1.2181$\pm$1.2744 & 0.0605$\pm$0.0120\\
LagLLama & 0.0001$\pm$0.0000 & 0.0897$\pm$0.0013 & 0.0590$\pm$0.0000 & 0.0062$\pm$0.0010 & \underline{0.0000$\pm$0.0000} & 0.0127$\pm$0.0007 \\

\bottomrule
\end{tabular}
\end{sc}
\end{scriptsize}
\end{center}
\vskip -0.1in
\end{table*}

\begin{table}[htbp]
\caption{Average Rank Across all Pre-Training Datasets. Lower is better.}
\label{table:training-performance-av-rank}
\vskip 0.15in
\begin{center}
\begin{scriptsize}
\begin{sc}
\setlength{\tabcolsep}{-0.6em} 
\begin{tabular}{lc}
\toprule
\multirow{2}{*}{\textbf{Model}} & \multirow{2}{*}{\textbf{Average Rank}} \\
 & \\
\midrule
ETSFormer & 10.900 \\
AutoETS & 10.200 \\
CrostonSBA & 10.000 \\
OFA & 9.850 \\
AutoFormer & 9.550 \\
NPTS & 8.350 \\
AutoARIMA & 8.333 \\
DynamicOptimize & 8.300 \\
Informer & 6.025 \\
DeepAR & 5.125 \\
PatchTST & 4.700 \\
LagLLama & 4.625 \\
NBEATS & 4.600 \\
TemporalFusionT & {3.875} \\
\bottomrule
\end{tabular}
\end{sc}
\end{scriptsize}
\end{center}
\vskip -0.1in
\end{table}

\section{Hyperparameters of \model} \label{app:lag-llama-hyperparameters}

We perform a random search of $100$ different hyperparameter configurations and use the average validation loss over all datasets in the pretraining corpus to select our model.

We list the possible hyperparameters of \model and the optimal values obtained by our hyperparameter search in \cref{tab:lag-llama-hps}. Our final model obtained by hyperparameter search contains 2,449,299 parameters.



\newpage
\newpage

\section{Forecast Visualizations} \label{app:forecast-visualizations}

\begin{figure*}[!htp]
    \centering
    \subfigure{\includegraphics[width=0.3\linewidth]{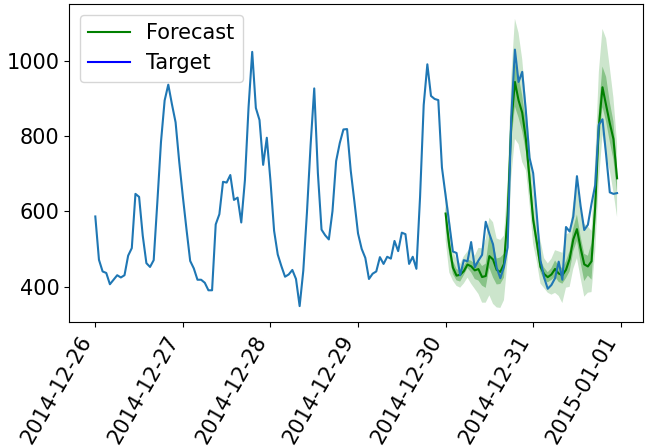}\label{fig:electricity-1}}
    \subfigure{\includegraphics[width=0.3\linewidth]{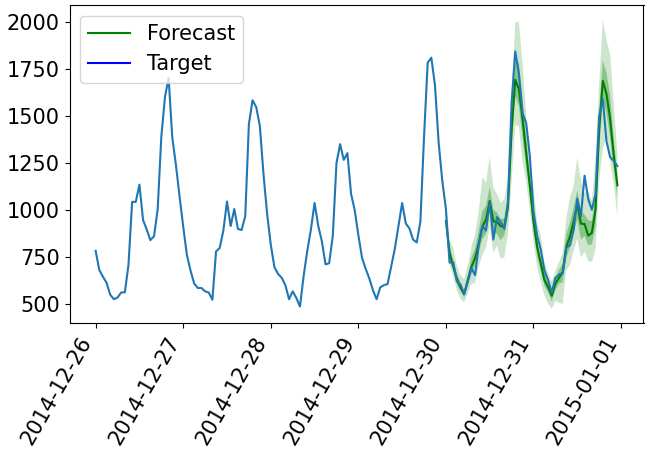}\label{fig:electricity-2}}
    \subfigure{\includegraphics[width=0.3\linewidth]{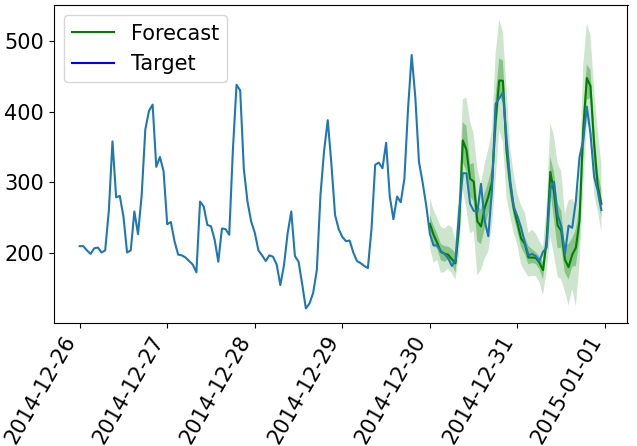}\label{fig:electricity-3}}
    \caption{Forecasting examples on the \texttt{Electricity Hourly} dataset}
    \label{fig:Pretrained forecast electricity}
\end{figure*}

\begin{figure*}[!htp]
    \centering
    \subfigure{\includegraphics[width=0.3\linewidth]{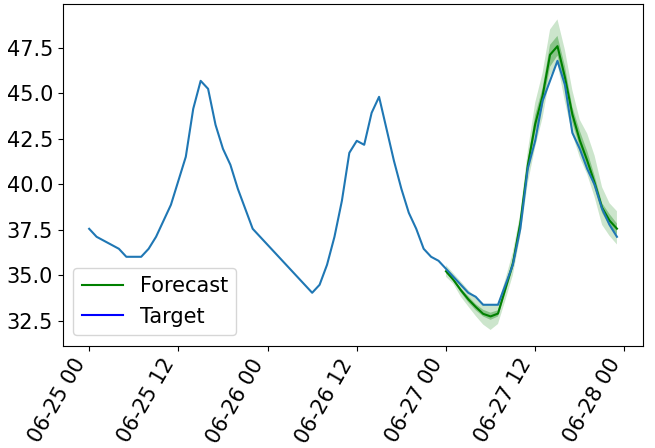}\label{fig:etth2-1}}
    \subfigure{\includegraphics[width=0.3\linewidth]{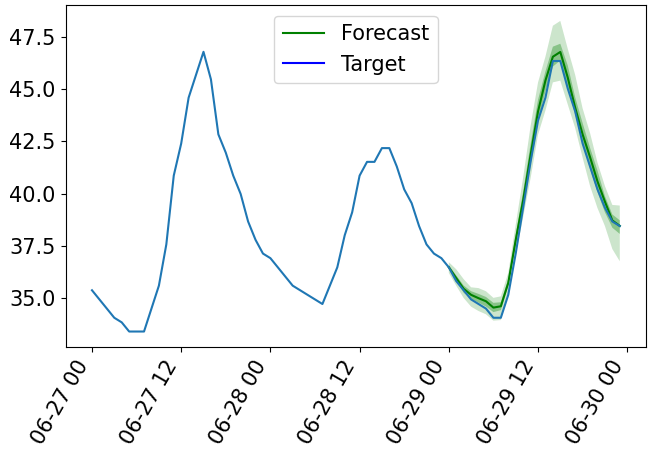}\label{fig:etth2-2}}
    \subfigure{\includegraphics[width=0.3\linewidth]{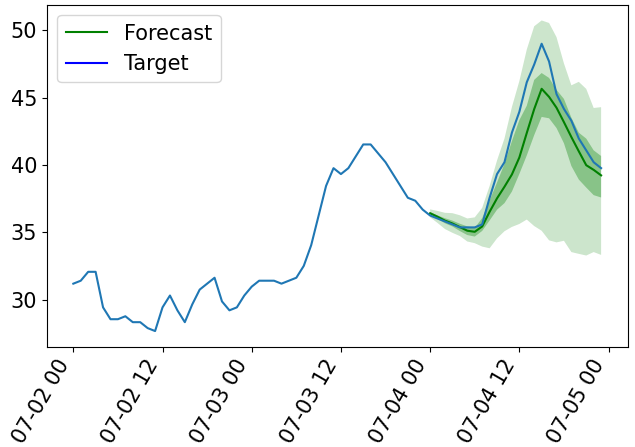}\label{fig:etth2-3}}
    \caption{Forecasting examples from \texttt{ETT-H2} dataset}
    \label{fig:Pretrained forecast etth2}
\end{figure*}

\begin{figure*}[!htp]
    \centering
    \subfigure{\includegraphics[width=0.3\linewidth]{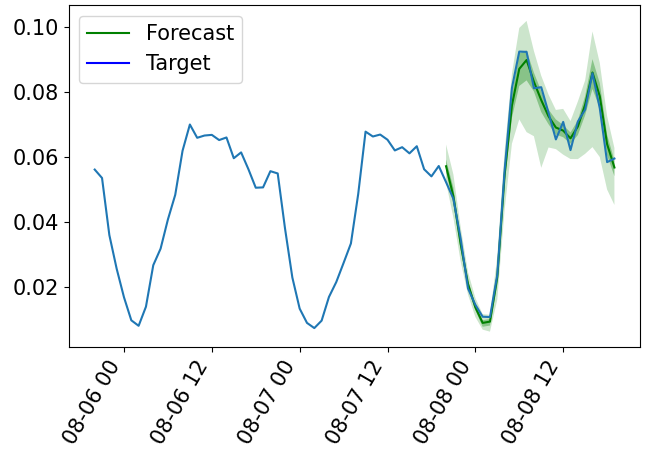}\label{fig:traffic-1}}
    \subfigure{\includegraphics[width=0.3\linewidth]{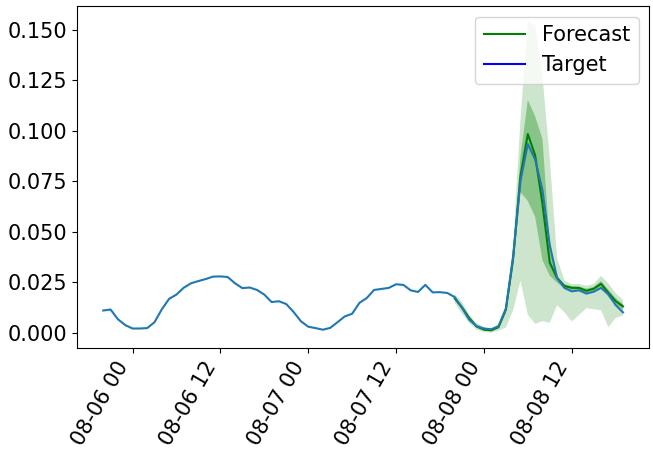}\label{fig:traffic-2}}
    \subfigure{\includegraphics[width=0.3\linewidth]{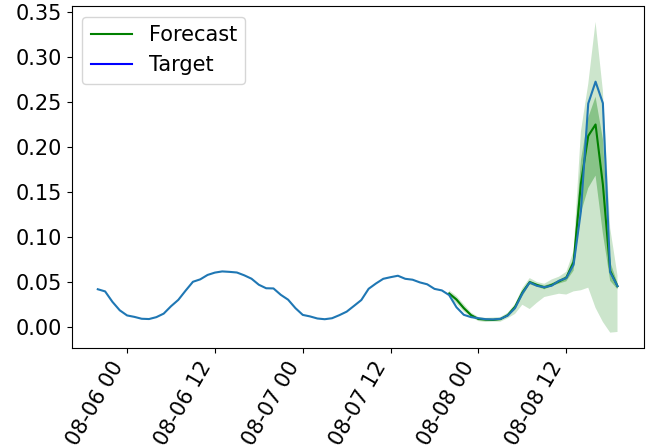}\label{fig:traffic-3}}
    \caption{Forecasting examples from \texttt{Traffic} dataset}
    \label{fig:Pretrained forecast traffic}
\end{figure*}

\begin{figure*}[!htp]
    \centering
    \subfigure{\includegraphics[width=0.3\linewidth]{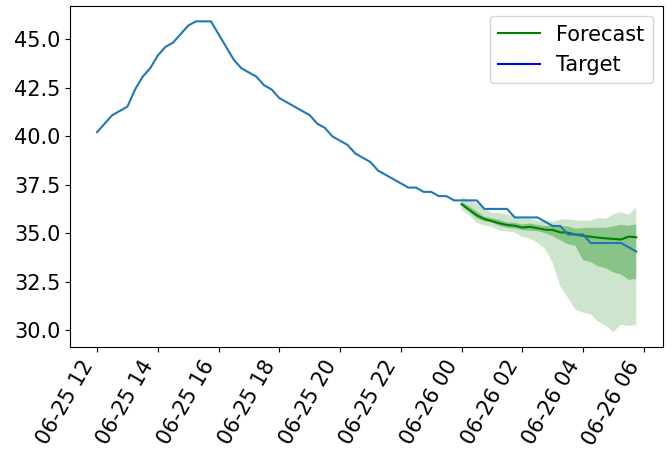}\label{fig:zeroshot-ettm2-1}}
    \subfigure{\includegraphics[width=0.3\linewidth]{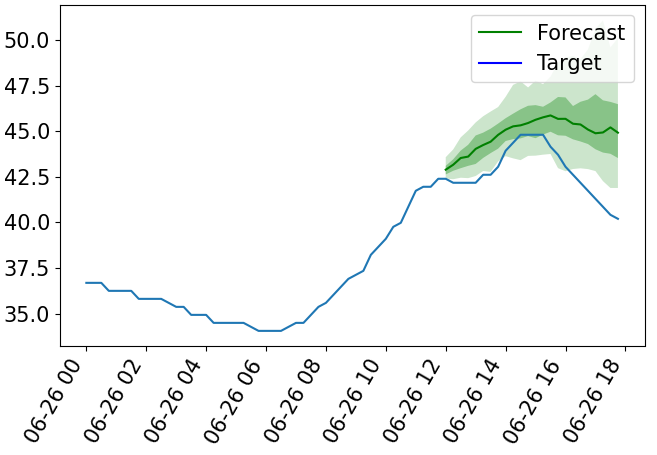}\label{fig:zeroshot-ettm2-2}}
    \subfigure{\includegraphics[width=0.3\linewidth]{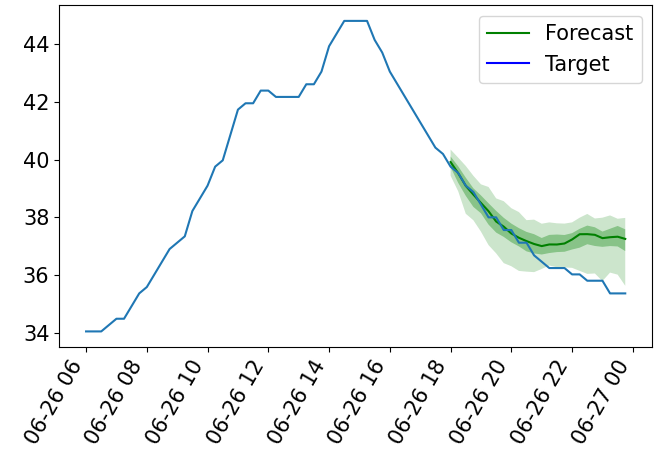}\label{fig:zeroshot-ettm2-3}}
    \caption{Zero-shot forecasting examples on the unseen downstream \texttt{ETT-M2} dataset}
    \label{fig:Zeroshot forecast ettm2}
\end{figure*}

\begin{figure*}[!htp]
    \centering
    \subfigure{\includegraphics[width=0.3\linewidth]{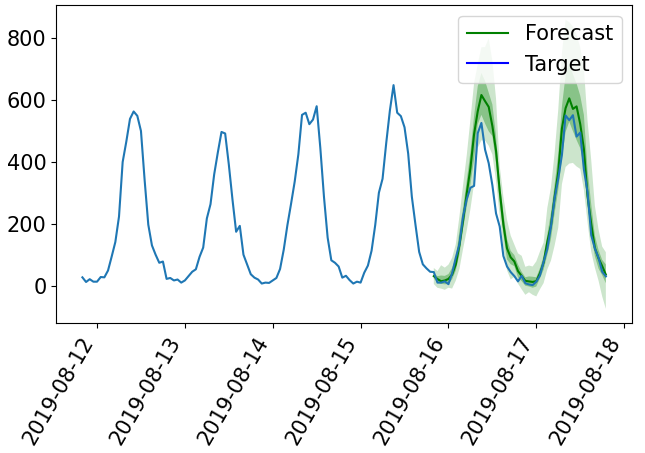}\label{fig:zeroshot-pedestrians-1}}
    \subfigure{\includegraphics[width=0.3\linewidth]{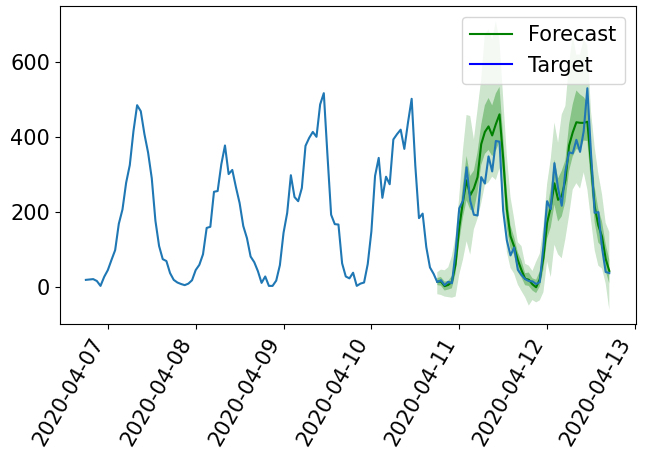}\label{fig:zeroshot-pedestrians-2}}
    \subfigure{\includegraphics[width=0.3\linewidth]{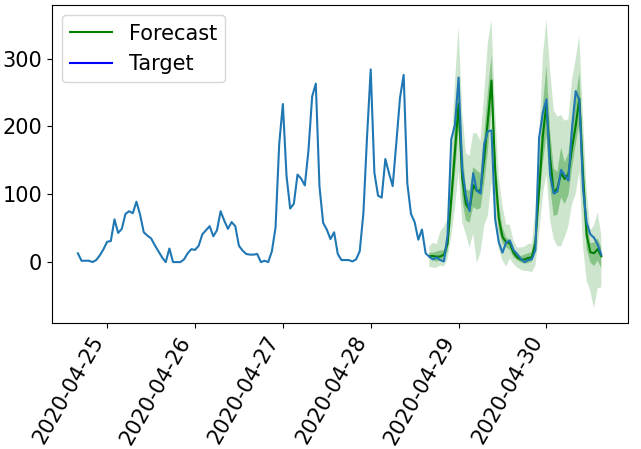}\label{fig:zeroshot-pedestrians-3}}
    \caption{Zero-shot forecasting examples on the unseen downstream \texttt{Pedestrian Counts} dataset}
    \label{fig:Zeroshot forecast pedestrians}
\end{figure*}

\begin{figure*}[!htp]
    \centering
    \subfigure{\includegraphics[width=0.3\linewidth]{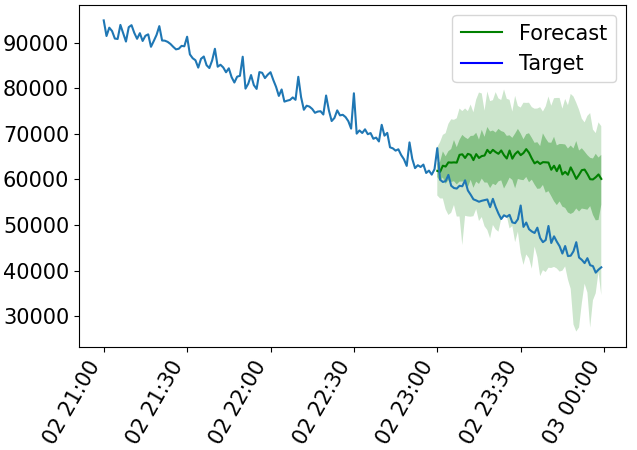}\label{fig:zeroshot-requests-1}}
    \subfigure{\includegraphics[width=0.3\linewidth]{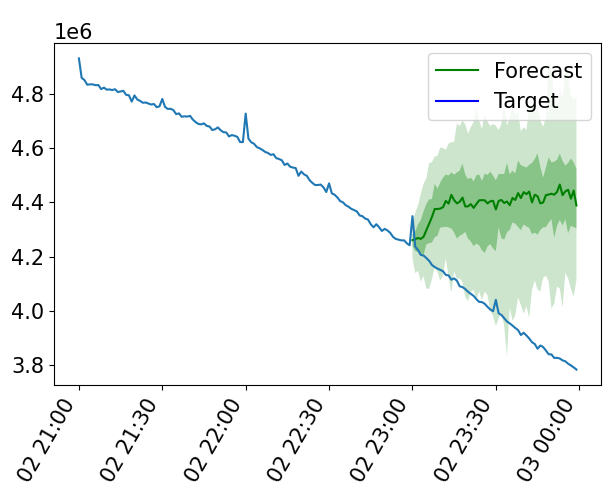}\label{fig:zeroshot-requests-2}}
    \subfigure{\includegraphics[width=0.3\linewidth]{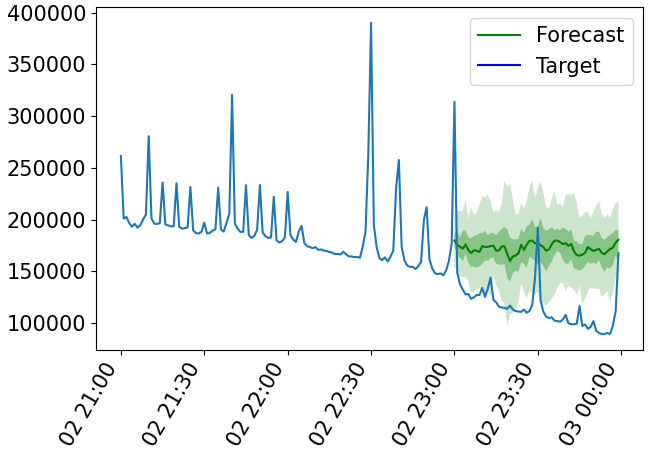}\label{fig:zeroshot-requests-3}}
    \caption{Zero-shot forecasting examples on the unseen downstream \texttt{Requests Minute} dataset}
    \label{fig:Zeroshot-forecast-requests}
\end{figure*}

\begin{figure*}[!htp]
    \centering
    \subfigure{\includegraphics[width=0.3\linewidth]{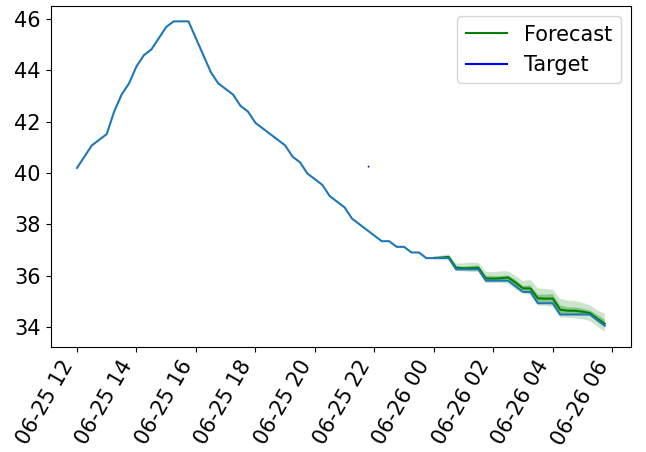}\label{fig:finetune-ettm2-1}}
    \subfigure{\includegraphics[width=0.3\linewidth]{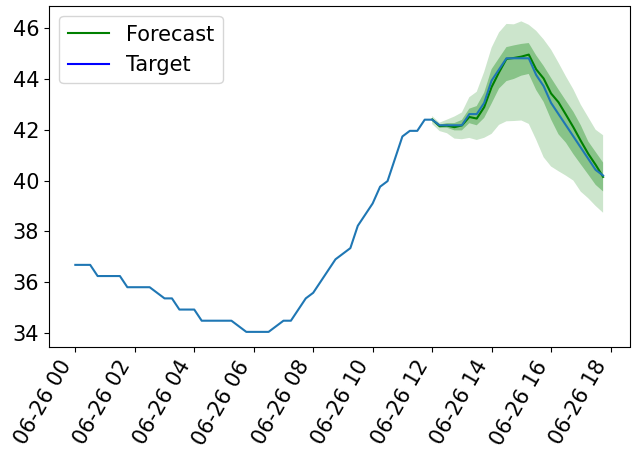}\label{fig:finetune-ettm2-2}}
    \subfigure{\includegraphics[width=0.3\linewidth]{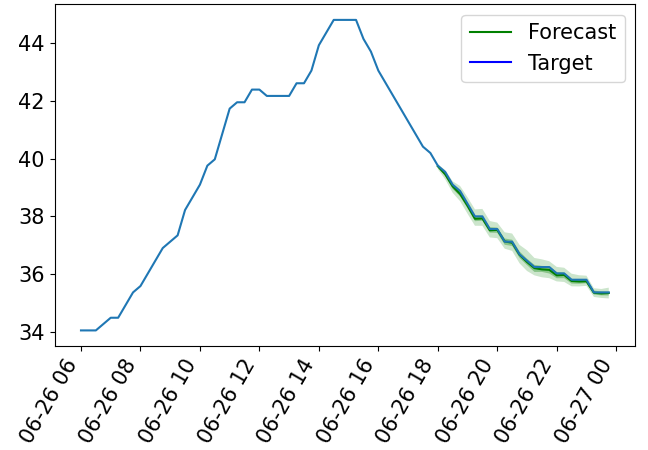}\label{fig:finetune-ttm2-3}}
    \caption{\model fine-tuned forecasting examples on the downstream \texttt{ETT-M2} dataset}
    \label{fig:FT forecast ettm2}
\end{figure*}

\begin{figure*}[!htp]
    \centering
    \subfigure{\includegraphics[width=0.3\linewidth]{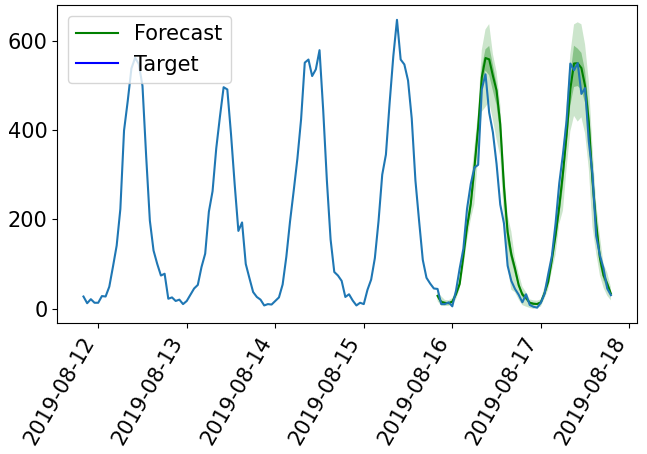}\label{fig:finetune-pedestrians-1}}
    \subfigure{\includegraphics[width=0.3\linewidth]{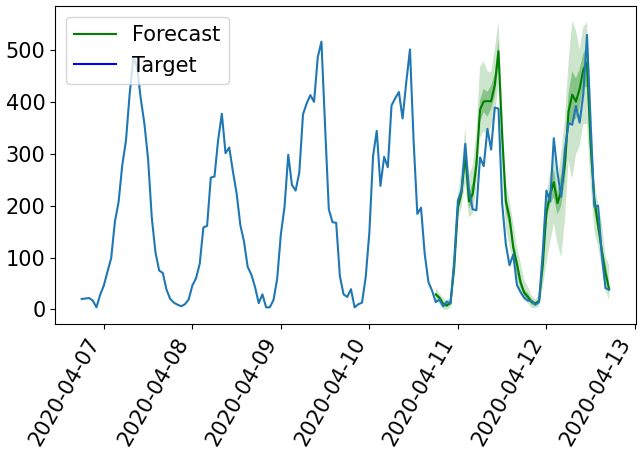}\label{fig:finetune-pedestrians-2}}
    \subfigure{\includegraphics[width=0.3\linewidth]{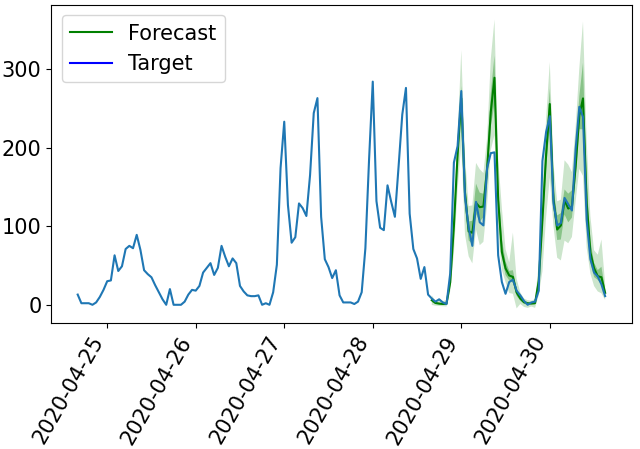}\label{fig:finetune-pedestrians-3}}
    \caption{\model fine-tuned forecasting examples on the downstream \texttt{Pedestrian Counts} dataset}
    \label{fig:FT forecast pedestrians}
\end{figure*}

\begin{figure*}[!htp]
    \centering
    \subfigure{\includegraphics[width=0.3\linewidth]{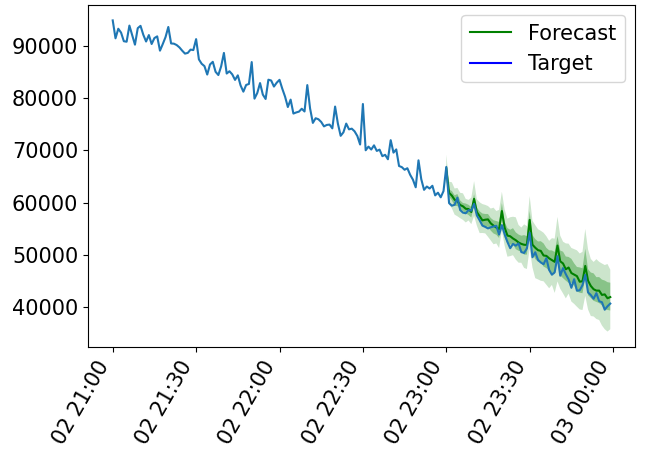}\label{fig:finetune-requests-1}}
    \subfigure{\includegraphics[width=0.3\linewidth]{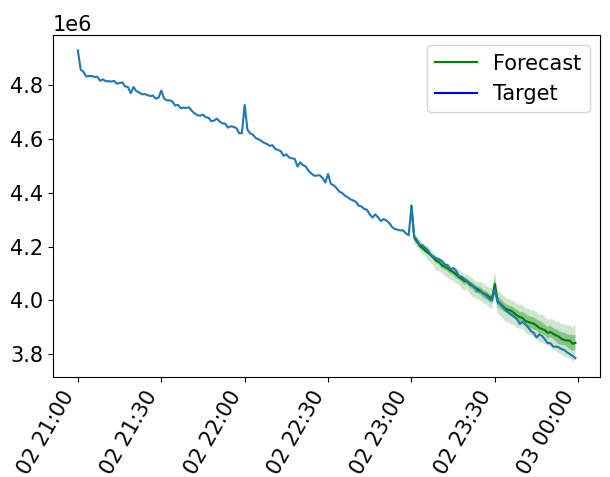}\label{fig:finetune-requests-2}}
    \subfigure{\includegraphics[width=0.3\linewidth]{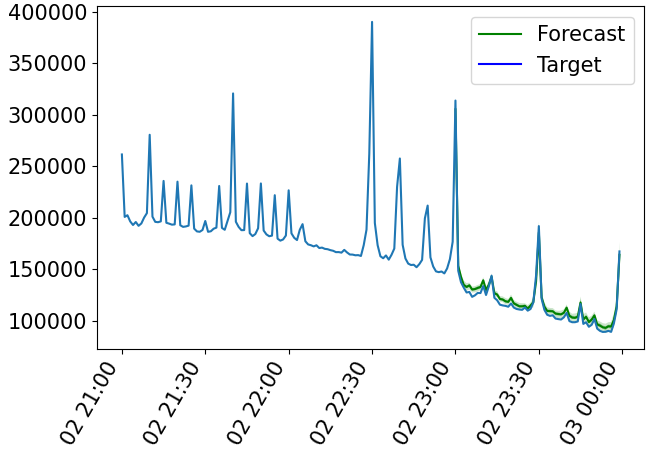}\label{fig:finetune-requests-3}}
    \caption{\model fine-tuned forecasting examples on the downstream \texttt{Requests Minute} dataset}
    \label{fig:FT-forecast-requests}
\end{figure*}

We plot some sample forecasts and highlight the median, 50-th (dark green) and 90-th (light-green) prediction interval; starting from datasets in the pretraining corpus: \texttt{Electricity Hourly} in Figure \ref{fig:Pretrained forecast electricity},  \texttt{ETT-H2} in Figure \ref{fig:Pretrained forecast etth2}, \texttt{Traffic} in Figure \ref{fig:Pretrained forecast traffic}. The Zero-shot forecasts of \model on downstream unseen datasets are highlighted for texttt{ETT-M2} in Figure \ref{fig:Zeroshot forecast ettm2},  \texttt{Pedestrian Counts} in Figure \ref{fig:Zeroshot forecast pedestrians} and \texttt{Requests Minute} in Figure \ref{fig:Zeroshot-forecast-requests}. Finally, forecasts after fine-tuning on these downstream unseen datasets are shown for \texttt{ETT-M2} in Figure \ref{fig:FT forecast ettm2}, \texttt{Pedestrian Counts} in Figure \ref{fig:FT forecast pedestrians} and \texttt{Requests Minute} in Figure \ref{fig:FT-forecast-requests}. Note in particular the different magnitudes of the sampled values depending on the dataset, via the same shared model.



\section{Additional Visualizations}\label{appendix:viz}


\begin{figure*}
    \centering
    \includegraphics[width=1\linewidth]{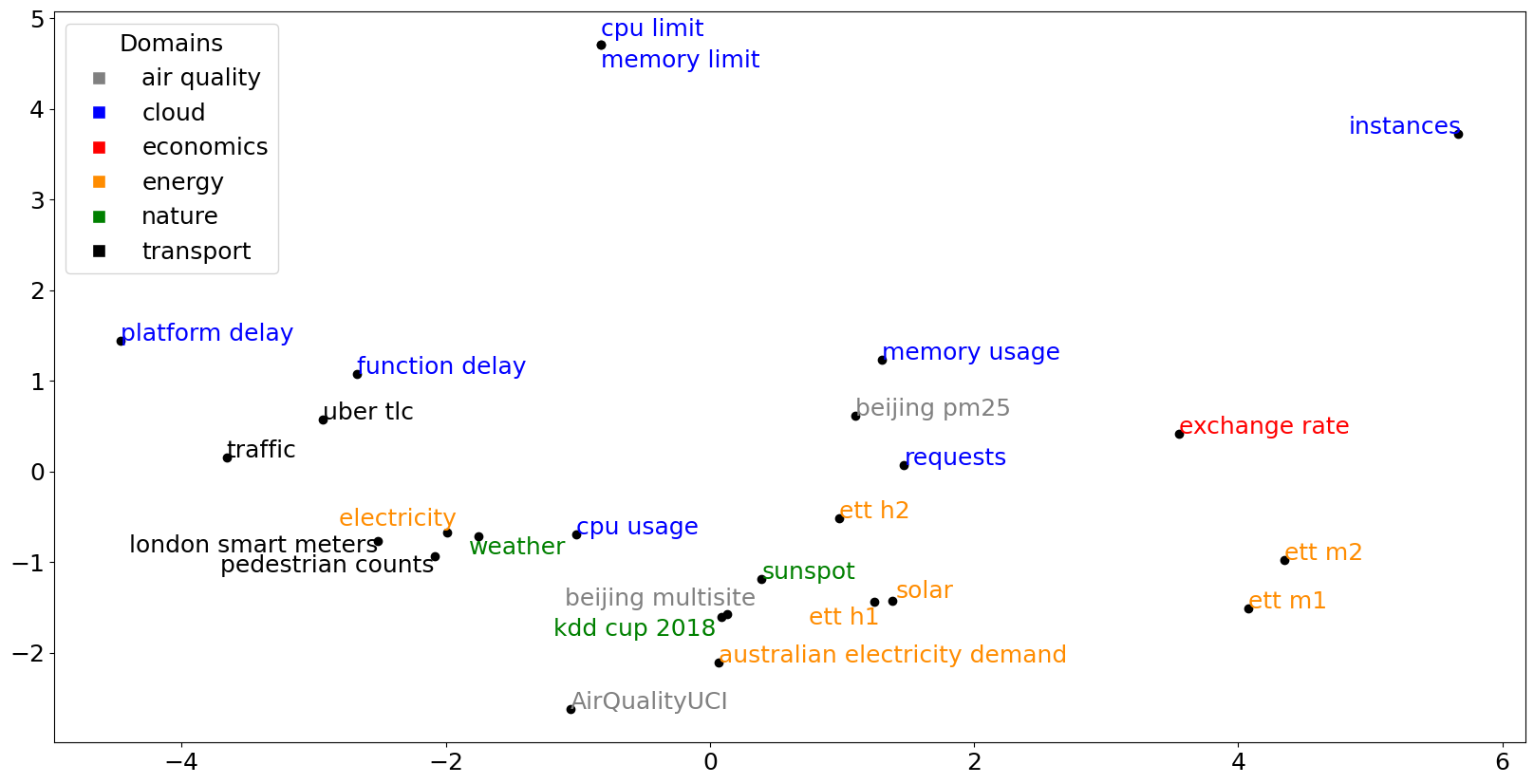}
    \caption{Principal Component Analysis (PCA) on the average catch22 features of each pre-training dataset.
    We take the average of the catch22 features for each dataset, standardize them, and then perform PCA on those points, such that each point corresponds to one dataset.
    We then visualize these points projected onto the top 2 components, and we color the name of each dataset according to its domain.
    The datasets are spread over both components, showing a diversity among the average catch-22 features of the different datasets.
    Also, datasets from different domains tend to be clustered together, which demonstrates that combining different domains increases pre-training data diversity.
    %
    %
    Together, these results suggest that combining multiple datasets across different domains increases the diversity of the pre-training data.
    Under the assumption that diversity in the pre-training data is beneficial for foundation model pre-training \cite{brown2020language}, pre-training a single time series model on a diverse combination of multiple datasets from multiple domains is beneficial to the foundation model's zero-shot and few-shot adaptation performance.
    }
    \label{fig:pca}
\end{figure*}


\subsection{Neural Scaling Laws} \label{app:neural-scaling-laws-contd}

\begin{figure*}[!htp]
    \centering
    \includegraphics[width=1\linewidth]{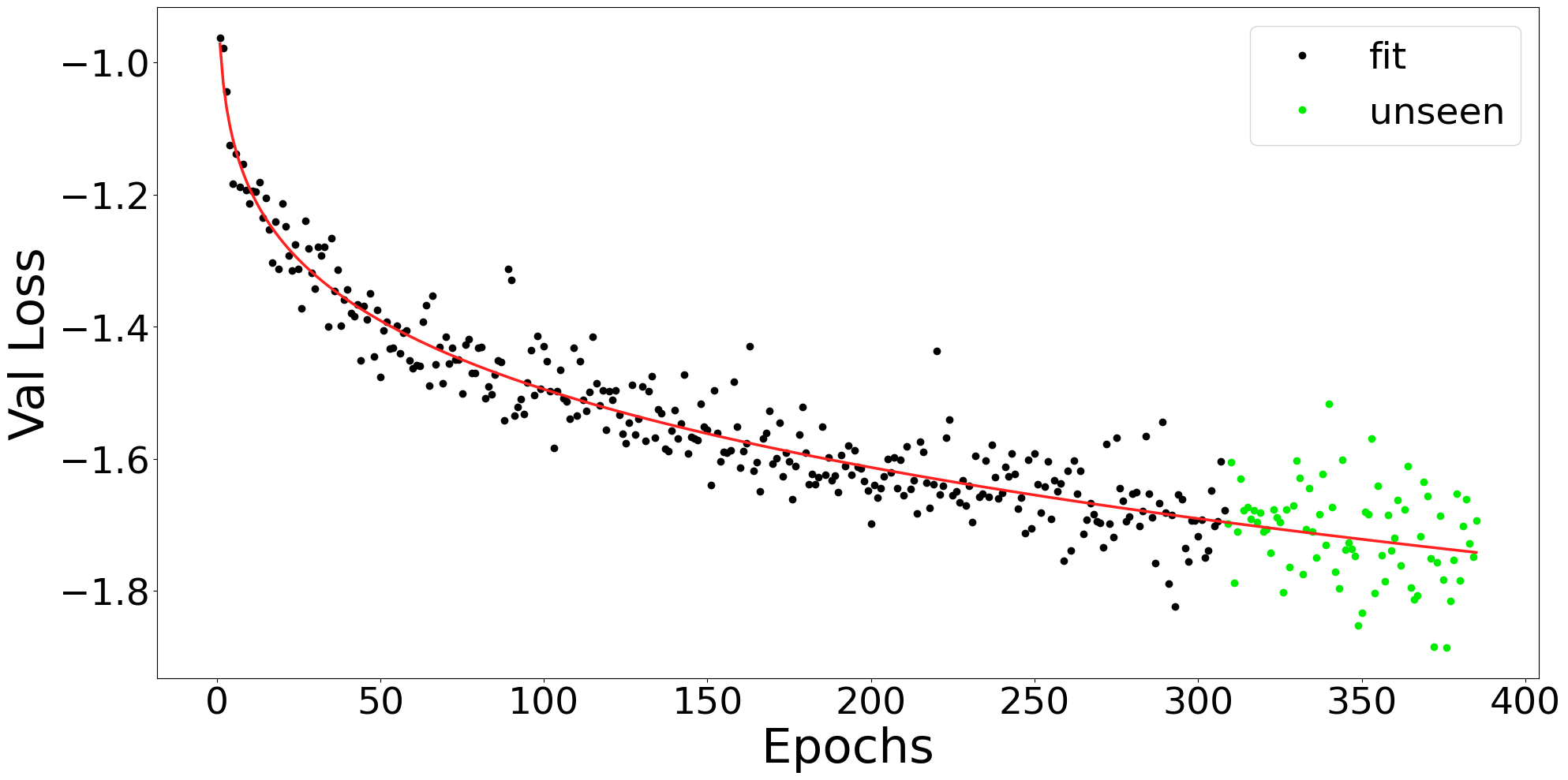}
    \caption{A neural scaling law fit to the validation loss (negative log-likelihood) of our foundation model, averaged across 3 seeds. "fit" represents points from the validation curve used for constructing the scaling law. "unseen" represents points of the validation curve that are predicted with the constructed scaling.
    We use a 60/20/20 train/val/test split to fit our scaling law.}
    \label{fig:bnsl}
\end{figure*}

The parameters of the Neural Scaling Law \citep{caballero2023broken} fit in Figure \ref{fig:bnsl} to the validation loss ($y)$ with respect to the pretraining data epochs seen ($x$) (where each epoch is $100$ randomly sampled windows) are given below. 
{\small
\begin{align*}
y           &=a+\left(b x^{-c_0}\right) \prod_{i=1}^n\left(1+\left(\frac{x}{d_i}\right)^{1 / f_i}\right)^{-c_i * f_i}\\
    a       &= -6.1167 \\
    b       &= 8.01589\\
    c_0     &= 0.0155\\
    c_1     &= -0.1043\\
    d_1     &= 1.6423e-36\\
    f_1     &= -36.4660 
\end{align*}
}

With such a law, one can extrapolate the validation loss of the model and predict performance in larger dataset regimes (Figure \ref{fig:bnsl}). As efforts progress towards collating better data repositories for time series foundation model training, such laws can help quantify the relations between the data used and the performance of the model.



\end{document}